\documentclass[11pt]{article}
\PassOptionsToPackage{table}{xcolor}
\usepackage{v2style}

\usepackage{amsmath,amsfonts,bm}

\def\eqref#1{equation~\ref{#1}}

\def\1{\bm{1}}

\DeclareMathAlphabet{\mathsfit}{\encodingdefault}{\sfdefault}{m}{sl}
\SetMathAlphabet{\mathsfit}{bold}{\encodingdefault}{\sfdefault}{bx}{n}

\usepackage{hyperref}
\usepackage{url}
\usepackage[all]{hypcap}
\usepackage{longtable,booktabs,array,calc,tabularx,xcolor}
\usepackage{graphicx}
\usepackage{placeins}
\usepackage{tikz}
\usepackage{amsmath}  %
\usepackage{algorithm}
\usepackage{algpseudocode}
\usepackage{wrapfig}
\providecommand{\tightlist}{\setlength{\itemsep}{0pt}\setlength{\parskip}{0pt}}
\usepackage{iftex}
\ifPDFTeX
  \usepackage[utf8]{inputenc}
  \usepackage[T1]{fontenc}
  \usepackage{textcomp}
  \newcommand{\paperunicode}[3]{\DeclareUnicodeCharacter{#1}{#3}}
\else
  \usepackage{fontspec}
  \usepackage{newunicodechar}
  \newcommand{\paperunicode}[3]{\newunicodechar{#2}{#3}}
\fi

\paperunicode{2265}{≥}{\ensuremath{\geq}}
\paperunicode{2264}{≤}{\ensuremath{\leq}}
\paperunicode{2248}{≈}{\ensuremath{\approx}}
\paperunicode{2260}{≠}{\ensuremath{\neq}}
\paperunicode{00D7}{×}{\ensuremath{\times}}
\paperunicode{2192}{→}{\ensuremath{\rightarrow}}
\paperunicode{2212}{−}{\ensuremath{-}}
\paperunicode{03C0}{π}{\ensuremath{\pi}}
\paperunicode{03B1}{α}{\ensuremath{\alpha}}
\paperunicode{03B2}{β}{\ensuremath{\beta}}
\paperunicode{03BB}{λ}{\ensuremath{\lambda}}
\paperunicode{0394}{Δ}{\ensuremath{\Delta}}
\paperunicode{03B8}{θ}{\ensuremath{\theta}}
\paperunicode{03C3}{σ}{\ensuremath{\sigma}}
\paperunicode{03BC}{μ}{\ensuremath{\mu}}
\paperunicode{226B}{≫}{\ensuremath{\gg}}
\paperunicode{226A}{≪}{\ensuremath{\ll}}
\paperunicode{00B1}{±}{\ensuremath{\pm}}
\paperunicode{00B7}{·}{\ensuremath{\cdot}}
\paperunicode{00B2}{²}{\ensuremath{^{2}}}
\paperunicode{2021}{‡}{\ensuremath{\ddagger}}
\paperunicode{2020}{†}{\ensuremath{\dagger}}
\paperunicode{21D2}{⇒}{\ensuremath{\Rightarrow}}
\paperunicode{2261}{≡}{\ensuremath{\equiv}}
\paperunicode{2153}{⅓}{\ensuremath{\tfrac{1}{3}}}
\paperunicode{03A3}{Σ}{\ensuremath{\Sigma}}
\paperunicode{03B3}{γ}{\ensuremath{\gamma}}
\paperunicode{03B5}{ε}{\ensuremath{\varepsilon}}
\paperunicode{03C4}{τ}{\ensuremath{\tau}}
\paperunicode{2190}{←}{\ensuremath{\leftarrow}}
\paperunicode{2208}{∈}{\ensuremath{\in}}
\paperunicode{221A}{√}{\ensuremath{\surd}}
\paperunicode{223C}{∼}{\ensuremath{\sim}}
\paperunicode{2026}{…}{\ldots{}}
\paperunicode{2713}{✓}{\ensuremath{\checkmark}}

\newsavebox{\findingbox}

\title{Beyond Teacher Assignment: Domain-Normalized\\ Multi-Teacher On-Policy Distillation}
\runningtitle{Beyond Teacher Assignment: Domain-Normalized Multi-Teacher On-Policy Distillation}
\author{Xin Li$^{1}$, Hao Jiang$^{1}$, Xin Gao$^{2}$, Annan Wang$^{1}$, Yuchen Xie$^{1}$,
Jinghao Guo$^{1}$, Xingwei Qu$^{3}$, Yichi Zhang and Chau Yuen$^{1}$}
\affiliations{$^{1}$Nanyang Technological University,\ $^{2}$Yale University,\
$^{3}$University of Manchester}
\projectpage{https://lixin.ai/DN-MOPD}
\titlelogos{\vtwologo{32pt}{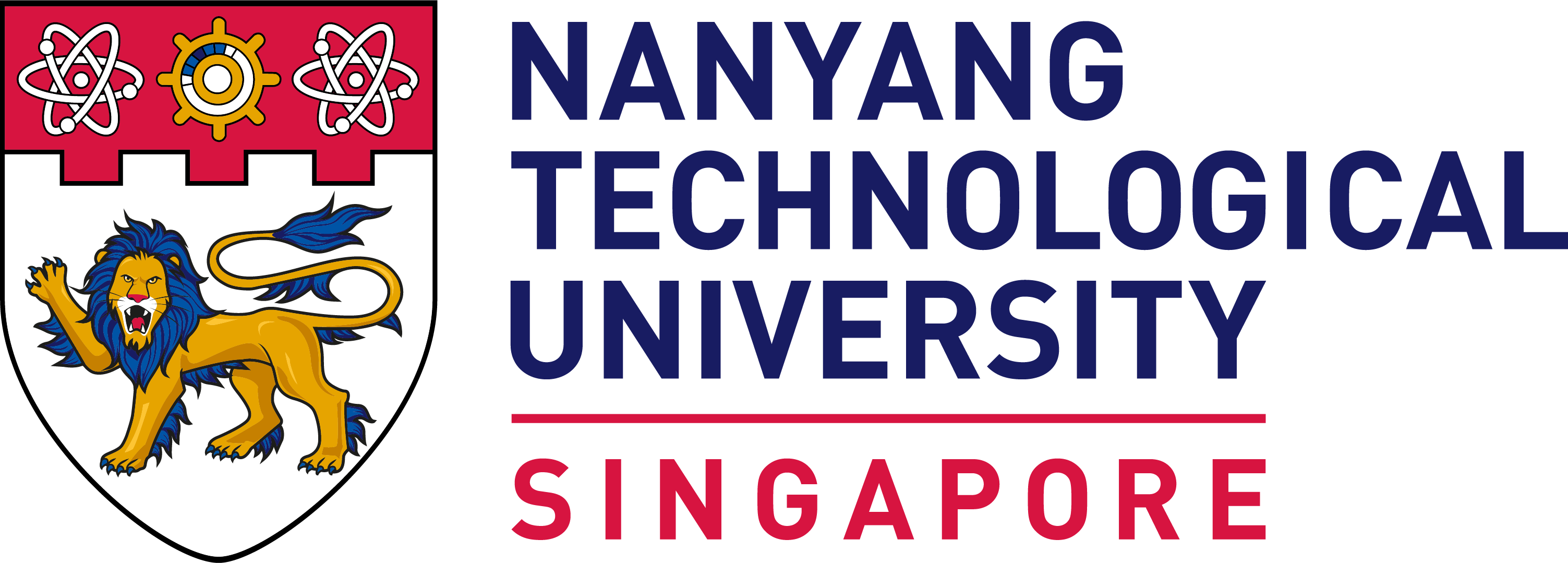}\vtwologosep
  \vtwologo{21pt}{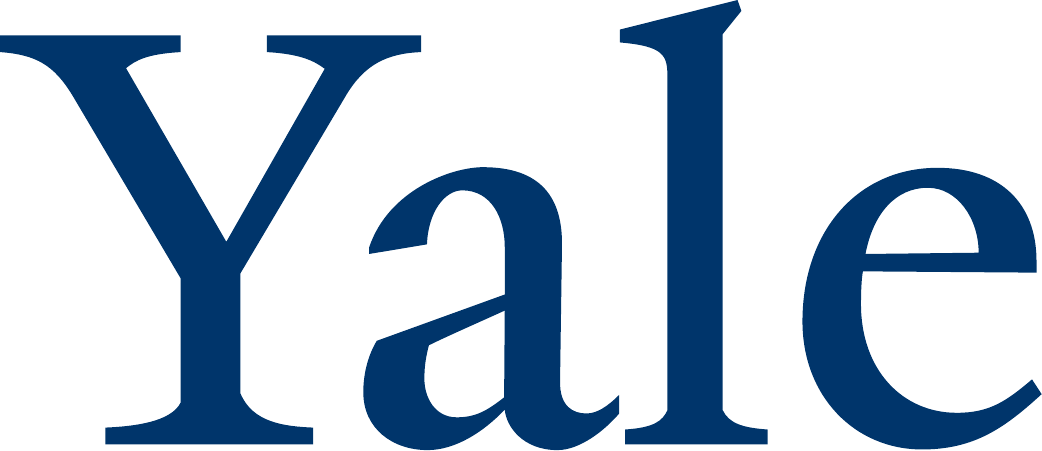}\vtwologosep
  \vtwologo{27pt}{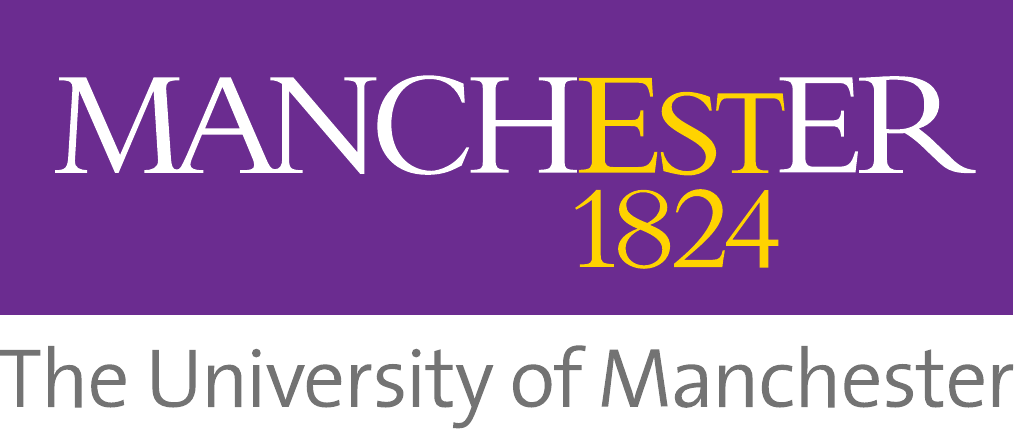}}
\date{September 2026}
\hypersetup{pdftitle={Beyond Teacher Assignment: Domain-Normalized Multi-Teacher On-Policy Distillation},
  pdfauthor={Xin Li, Hao Jiang, Xin Gao, Annan Wang, Yuchen Xie, Jinghao Guo, Xingwei Qu, Yichi Zhang, Chau Yuen}}
\begin{document}
\maketitle
\begin{abstract}
Reinforcement learning can turn one language model into several
specialists, each excellent at a single skill such as mathematics,
coding or following instructions, but users need one model with all of
these skills. Multi-teacher on-policy distillation (MOPD) merges them by
letting the specialists teach one student: the student answers each
prompt, and the specialist for that prompt's domain gives feedback on
every token. This routing decides which specialist teaches, but not how
strongly its feedback moves the shared student. In Qwen3.5 models at
three sizes, we find that MOPD's student does not beat one taught by the
best single specialist and gains little of the mathematics specialist's
advantage. The feedback is unbalanced: instruction-following feedback is
several times more spread out than mathematics feedback and dominates
the student's updates. We propose Domain-Normalized MOPD (DN-MOPD),
which keeps the routing and rescales each domain's feedback by its
measured spread. On six public benchmarks, DN-MOPD improves the average
score over MOPD at every size, across three random seeds and under two
answer-length limits, and recovers most of the lost mathematics gain.
Controls with fixed domain weights show that the gain comes mainly from
turning down instruction-following feedback rather than turning up
mathematics alone, and that fixed weights close to those DN-MOPD
measures perform comparably. Combining specialists therefore requires
deciding not only which one teaches, but also how strongly its feedback
counts.

\end{abstract}
\section{Introduction}\label{sec:1}\label{introduction}

Post-training of large language models increasingly relies on
domain-specific reinforcement learning (RL), with verifiable rewards for
mathematical reasoning \citep{arxiv240203300,arxiv250112948}, execution
feedback for code \citep{arxiv250201456}, and constraint checking for
instruction following \citep{arxiv241115124,arxiv250702833}. Because
these pipelines differ in data, rewards and optimization, they are
usually run separately from a shared initialization, yielding experts
that each excel in one domain. A deployable model must integrate them
\citep{arxiv260630406}.

\begin{figure}[ht]\centering\includegraphics[width=.8\linewidth]{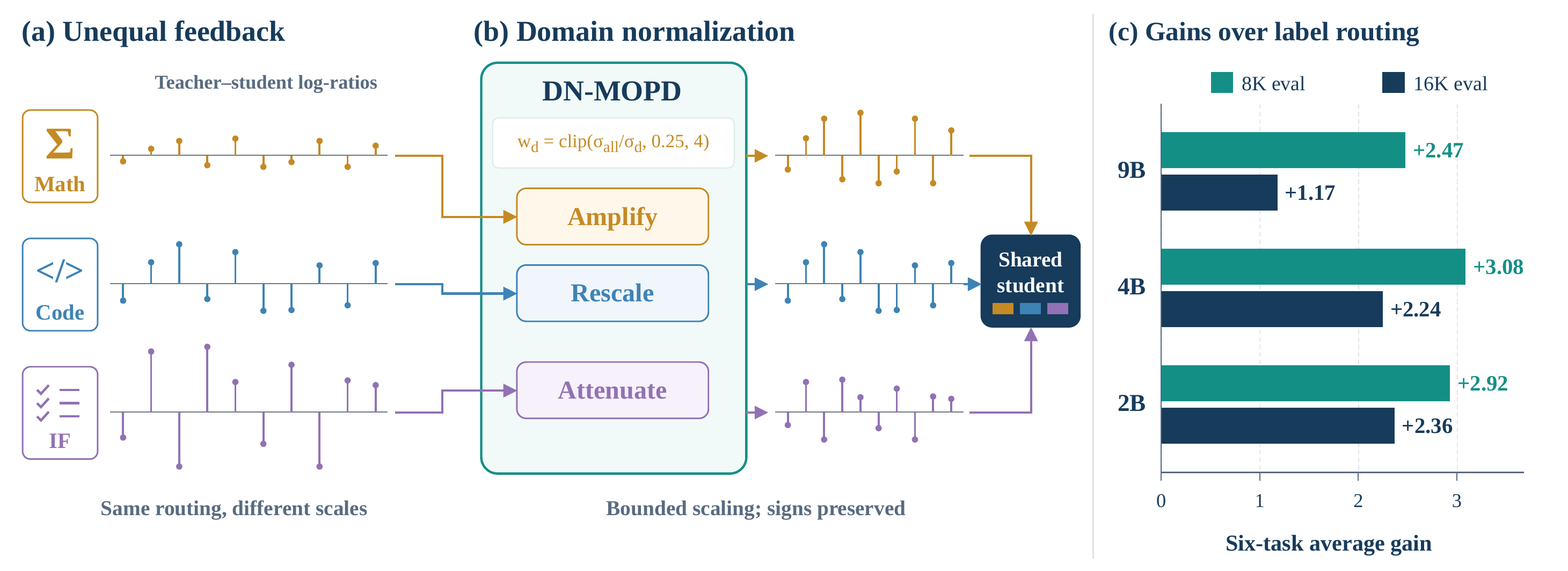}\caption{Label routing decides which teacher supervises a prompt; DN-MOPD also controls how strongly each teacher's feedback counts. (a) Teacher–student log-ratios differ in scale across domains (schematic). (b) DN-MOPD rescales each domain's feedback with a bounded, sign-preserving multiplier (schematic). (c) Six-task Total gain of DN-MOPD over label routing; paired 95\% intervals are reported in \hyperref[app:C]{App. C}.}\label{fig:1}\end{figure}

\vspace{-4pt}

Multi-teacher on-policy distillation (MOPD) performs this integration in
policy space \citep{arxiv260630406,arxiv260102780}: the student samples
responses, each prompt is routed to the expert of its domain, and that
expert's token-level log-probabilities on the student's own response
provide a dense distillation signal
\citep{arxiv230613649,arxiv230608543}. MOPD has been adopted in the
post-training of several recent frontier models
\citep{arxiv260619348,deepseek2026v41,arxiv260102780,arxiv260615007,arxiv260720062,arxiv260809119},
and follow-up work reweights each domain's loss by its token share and
its remaining teacher--student gap \citep{arxiv260819098}. The routing
rule, however, specifies only which expert supervises a prompt, and
existing weights are fixed by hand or set from the size of the remaining
gap. Neither accounts for the spread of each expert's feedback, which
differs when experts are trained by different RL pipelines.

Our empirical investigation reveals that this choice matters: across
independently trained Qwen3.5 expert pools at 9B, 4B and 2B, MOPD with
label routing (Label in our tables) does not outperform the strongest
single-teacher student at any size and transfers little of the
mathematics expert's gain. At an 8K evaluation budget the student
retains only 14--32\% of that gain, and at 16K it shows no gain over its
initialization. The feedback itself is unbalanced. Token-level
teacher--student log-ratios from the instruction-following expert are
2.3 to 4.4 times as dispersed as the pooled signal of the first training
batch, whereas mathematics feedback is about half as dispersed (\hyperref[app:D.2]{App.
D.2}). Because all domains update the same parameters, this imbalance
shapes the update even when prompt counts are equal: for the initial 4B
student, the instruction-following loss supplies 94\% of the combined
gradient under equal weights (\hyperref[app:D.5]{App. D.5}). Routing determines where
feedback comes from, but not how much it counts.

We introduce \textbf{Domain-Normalized Multi-Teacher On-Policy
Distillation (DN-MOPD)}. For each batch, it measures the spread of
teacher--student log-ratios within each domain and uses a bounded
multiplier to bring that domain's distillation advantages toward the
pooled scale. The selected experts continue to supervise fresh student
answers. This adds one operation to MOPD, with no additional teacher
model, teacher call or learned router; MOPD is the special case in which
every multiplier equals one.

Across independently trained Qwen3.5 expert pools at 9B, 4B and 2B,
DN-MOPD improves the six-task average over MOPD at both evaluation
budgets (\hyperref[fig:1]{Figure 1}): by 1.17 to 2.36 percentage points at 16K and 2.47 to
3.08 at 8K. The advantage holds across three student seeds, and DN-MOPD
also exceeds the strongest single-teacher student at every size, with
some of these intervals including zero. Mathematics, which MOPD failed
to transfer, carries the largest gains. Controls with fixed domain
weights show that most of them come from limiting the
instruction-following feedback rather than from amplifying mathematics
alone, and that fixed weights near DN-MOPD's measured multipliers
perform comparably at 9B and 4B; DN-MOPD obtains this calibration from
batch statistics without a per-size weight search.

Our contributions are:

\begin{itemize}
\tightlist
\item
  \textbf{Teacher assignment alone does not transfer the specialists'
  skills.} MOPD with label routing does not outperform the strongest
  single-teacher student at any of three Qwen3.5 sizes and transfers
  little of the mathematics expert's gain. Its domains give feedback on
  unequal scales: in the first training batch, instruction-following
  log-ratios are 2.3--4.4 times as dispersed as the pooled signal and
  mathematics log-ratios about half as dispersed, and for the initial 4B
  student the instruction-following loss supplies 94\% of the combined
  gradient.
\item
  \textbf{DN-MOPD puts every teacher's feedback on a common scale.} It
  rescales each domain's distillation advantages by the clipped ratio of
  pooled to domain log-ratio spread, estimated on every batch, while
  keeping label routing and the sign of every advantage; MOPD is the
  special case in which every multiplier is one.
\item
  \textbf{Calibrating feedback scale consistently improves on MOPD.}
  DN-MOPD improves the six-task average over MOPD at every size and both
  evaluation budgets, with every paired interval above zero and the gain
  holding across three student seeds; it also exceeds the strongest
  single-teacher student and recovers most of the mathematics gain that
  MOPD loses.
\end{itemize}

\section{DN-MOPD}\label{sec:2}\label{dn-mopd}

\begin{figure}[t]\centering\includegraphics[width=.8\linewidth]{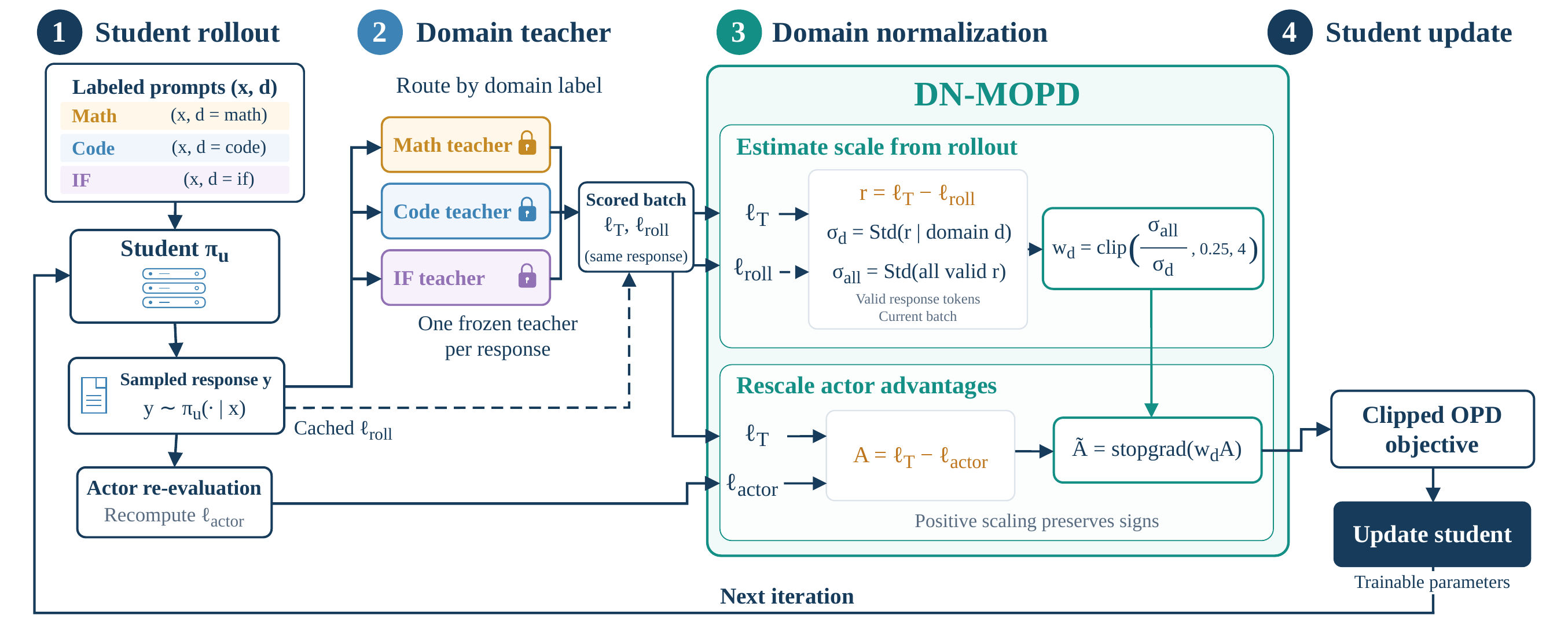}\caption{One DN-MOPD training iteration. (1) The student samples responses to labeled prompts. (2) Each response is scored by the frozen teacher of its domain. (3) Each domain's multiplier is estimated from cached rollout log-ratios and applied to the actor-side advantages. (4) The student is updated with the clipped OPD objective. Label routing is the special case in which every multiplier is 1.}\label{fig:2}\end{figure}

DN-MOPD keeps the domain-label assignment of multi-teacher OPD and adds
one operation: before the shared student is updated, each domain's
distillation advantages are rescaled using its feedback spread relative
to the current batch (\hyperref[fig:2]{Figure 2}). We build on the multi-teacher OPD
recipe of \citet{arxiv260630406} and the Uni-OPD implementation
\citep{arxiv260503677}.

\textbf{Setting.} We start from one initial model. Three copies are
trained separately with reinforcement learning into specialists for
mathematics, code and instruction following (IF). A fourth copy, the
student \(\pi_u\), is trained to acquire all three skills. Every
training prompt \(x\) carries a domain label
\(d\in\{\text{math},\text{code},\text{IF}\}\), and \(T_d\) denotes the
specialist for that domain.

\textbf{How a teacher gives feedback.} The student first writes its own
answer \(y=(y_1,\ldots,y_n)\) to the prompt. The matching teacher then
reads the same answer and, at every position \(t\), reports how likely
it would have been to write the student's token given the preceding text
\(h_t=(x,y_{<t})\). Comparing the two probabilities gives the token's
distillation advantage \[
A_t=\log p_{T_d}(y_t\mid h_t)-\log\pi_u(y_t\mid h_t).
\] A positive \(A_t\) encourages the sampled token, while a negative
\(A_t\) discourages it. Its magnitude weights that token's
policy-gradient contribution. Label-routed OPD uses \(A_t\) directly.

\textbf{Why the size of feedback matters.} All domains update the same
student parameters, but independently trained specialists need not
produce advantages on comparable scales. Domain labels select the source
of supervision without calibrating these scalar weights. In the first
training batch of our Qwen3.5 runs, IF log-ratios are 2.3--4.4 times as
dispersed as the pooled batch signal, and mathematics log-ratios about
half as dispersed (\hyperref[app:D.2]{App. D.2}). This motivates an explicit adjustment of
feedback scale. The scale affects the weighting of token contributions;
the resulting domain gradient also depends on their directions and the
loss reduction.

\textbf{Domain normalization.} We use standard deviation to measure the
dispersion of the feedback within each domain. For response \(i\), let
\(r_{i,t}=\ell^T_{i,t}-\ell^{\mathrm{roll}}_{i,t}\), where \(\ell^T\) is
the assigned teacher's token log-probability and
\(\ell^{\mathrm{roll}}\) is the student's cached rollout
log-probability. In each batch, \(\sigma_d\) is the population standard
deviation of \(r_{i,t}\) over valid response tokens with \(d_i=d\). We
compute \(\sigma_{\mathrm{all}}\) over all valid response tokens pooled
across domains. Each domain receives one multiplier: \[
w_d=\operatorname{clip}\!\left(\frac{\sigma_{\mathrm{all}}}{\sigma_d},\,0.25,\,4\right),
\qquad
\widetilde A_t=w_d A_t.
\] The pooled spread supplies a common reference that follows the
current batch. For a nondegenerate domain whose ratio is not clipped,
the scaled rollout signal satisfies \[
\operatorname{Std}(w_d r_d)=\sigma_{\mathrm{all}}.
\] Thus the rule amplifies feedback with a smaller spread and attenuates
feedback with a larger spread. It targets dispersion rather than
treating the statistic as an estimate of teacher quality or remaining
capability gap. This differs from allocating more budget to domains with
larger mean absolute rewards, as in Open-MOPD \citep{arxiv260819098}.

We multiply by a positive factor without subtracting the domain mean,
preserving each advantage's sign and hence whether it encourages or
discourages the sampled token. The mean is scaled along with the rest of
the signal. The bounds \([0.25,4]\) limit the adjustment when a scale
ratio is extreme; clipping can prevent full equalization. We use
\(w_d=1\) if either standard deviation is zero or has fewer than two
observations. Matching these spreads does not fix the batch loss scale
or equalize domain gradient norms.

\textbf{Student update.} The actor recomputes the pre-update student
log-probabilities \(\ell^{\mathrm{actor}}\) on the sampled responses,
forming \(A_{i,t}=\ell^T_{i,t}-\ell^{\mathrm{actor}}_{i,t}\). We detach
the scaled advantage,
\(\widetilde A_{i,t}=\operatorname{stopgrad}(w_{d_i}A_{i,t})\), and use
it in the baseline clipped OPD loss. With
\(\rho_{i,t}(\theta)=\pi_\theta(y_{i,t}\mid h_{i,t})/\exp(\ell^{\mathrm{actor}}_{i,t})\),
the token loss is \[
\mathcal L_{i,t}(\theta)=-\min\!\left\{
\rho_{i,t}(\theta)\widetilde A_{i,t},\;
\operatorname{clip}\bigl(\rho_{i,t}(\theta),1-\eta_-,1+\eta_+\bigr)\widetilde A_{i,t}
\right\},
\] where \(\eta_-\) and \(\eta_+\) are the baseline policy-ratio
clipping bounds. We retain Label's response masks and loss reduction:
average valid token losses within each response, then average across
responses. Scale estimation pools tokens, whereas the training loss
averages responses. Setting every \(w_d\) to one recovers Label.
Algorithm 1 summarizes the iteration; \hyperref[app:D.1]{App. D.1} gives the implementation
details.

\begin{wrapfigure}{r}{0.5\textwidth}
\vspace{-16pt}
\begin{minipage}{\linewidth}
\begin{algorithm}[H]
\footnotesize
\caption{DN-MOPD, one training iteration}\label{alg:dnmopd}
\begin{algorithmic}[1]
\Require student $\pi_u$, teachers $\{T_d\}$, labeled prompts
\State Sample responses $y_i\sim\pi_u(\cdot\mid x_i)$
\State Score each $y_i$ with its domain teacher $T_{d_i}$
\State $r_{i,t}\gets\ell^T_{i,t}-\ell^{\mathrm{roll}}_{i,t}$ on valid tokens
\State $\sigma_{\mathrm{all}}\gets\mathrm{Std}(\{r_{i,t}\})$
\For{each domain $d$}
\State $\sigma_d\gets\mathrm{Std}(\{r_{i,t}:d_i=d\})$
\State $w_d\gets\mathrm{clip}(\sigma_{\mathrm{all}}/\sigma_d,\,0.25,\,4)$
\State Use $w_d=1$ if either statistic is degenerate
\EndFor
\State Recompute $A_{i,t}\gets\ell^T_{i,t}-\ell^{\mathrm{actor}}_{i,t}$
\State $\widetilde A_{i,t}\gets\mathrm{stopgrad}(w_{d_i}A_{i,t})$
\State Update $\pi_u$ on $\widetilde A$ with the clipped OPD objective
\end{algorithmic}
\end{algorithm}
\end{minipage}
\vspace{-14pt}
\end{wrapfigure}

\textbf{What the rule does in practice.} In the Qwen3.5 runs, the
multiplier amplifies mathematics feedback by about 1.5--2.1 at every
size and keeps code near 1 at 4B and 2B. The IF multiplier sits at the
0.25 floor in most batches, so clipping bounds rather than equalizes
that domain's scale (\hyperref[fig:4]{Figure 4}).

\textbf{Cost and scope.} The operation needs no additional teacher call:
it reuses the log-ratios already available from rollout scoring,
computes one pooled standard deviation and one per domain, and scales
the existing advantages. It changes neither which teacher supervises a
prompt nor how many prompts each domain receives.

\section{Experiments}\label{sec:3}\label{experiments}

We first compare DN-MOPD with label-routed OPD, then examine which
specialist capabilities improve and how these changes relate to the
feedback scales. We also test sensitivity to evaluation length and
training duration.

\begin{table}[t]\centering
\caption{\textbf{Capability integration at 9B.} Qwen3.5-9B scores (\%) on six public tasks. Total averages the six task scores.}\label{tab:main}
\definecolor{qgray}{RGB}{244,245,246}
\definecolor{qblue}{RGB}{233,242,249}
\begingroup\fontsize{9}{10.5}\selectfont\renewcommand{\arraystretch}{1.03}\setlength{\tabcolsep}{3pt}
\begin{tabularx}{\linewidth}{@{}>{\raggedright\arraybackslash}Xrrrrrrr@{}}
\toprule
& \multicolumn{2}{c}{Math (avg@64)} & \multicolumn{2}{c}{Code (avg@6)} & \multicolumn{2}{c}{IF (avg@16)} & \\
\cmidrule(lr){2-3}\cmidrule(lr){4-5}\cmidrule(lr){6-7}
Method & AIME25 & AIME26 & LCB v5 & LCB v6 & IFEval & IFBench & Total \\
\midrule
Initial student & 57.7 & 62.6 & 54.9 & 51.4 & 82.4 & 33.8 & 57.1 \\
\addlinespace[2pt]\rowcolor{qgray}\multicolumn{8}{@{}l@{}}{\textit{RL experts}} \\
\leftskip=1em\relax Math expert & 59.7 & 69.4 & 53.4 & 49.4 & 82.3 & 34.9 & 58.2 \\
\leftskip=1em\relax Code expert & 58.5 & 65.4 & 58.5 & 54.3 & 82.3 & 34.6 & 58.9 \\
\leftskip=1em\relax IF expert & 54.6 & 60.5 & 54.0 & 51.5 & 86.5 & 42.1 & 58.2 \\
\addlinespace[2pt]\rowcolor{qgray}\multicolumn{8}{@{}l@{}}{\textit{Offline distillation}} \\
\leftskip=1em\relax SeqKD-SFT & 61.3 & 67.6 & 62.7 & 56.0 & 86.0 & 42.4 & 62.6 \\
\addlinespace[2pt]\rowcolor{qgray}\multicolumn{8}{@{}l@{}}{\textit{Parameter merging}} \\
\leftskip=1em\relax ParamMerge-Avg & 59.4 & 68.3 & 54.0 & 50.9 & 84.1 & 36.4 & 58.9 \\
\leftskip=1em\relax ParamMerge-TA & 59.1 & 68.6 & 55.2 & 50.9 & 86.1 & 43.3 & 60.5 \\
\addlinespace[2pt]\rowcolor{qgray}\multicolumn{8}{@{}l@{}}{\textit{Single-teacher OPD}} \\
\leftskip=1em\relax Math teacher & 58.9 & 69.5 & 53.7 & 49.4 & 82.4 & 34.7 & 58.1 \\
\leftskip=1em\relax Code teacher & 58.3 & 66.2 & 57.0 & 52.6 & 82.6 & 34.2 & 58.5 \\
\leftskip=1em\relax IF teacher & 56.8 & 61.9 & 52.1 & 49.2 & 84.7 & 38.9 & 57.3 \\
\addlinespace[2pt]\rowcolor{qgray}\multicolumn{8}{@{}l@{}}{\textit{Multi-teacher OPD}} \\
\leftskip=1em\relax Uniform pool & 57.1 & 65.9 & 53.6 & 52.3 & 83.3 & 35.7 & 58.0 \\
\leftskip=1em\relax Dynamic router & 56.4 & 63.5 & 55.4 & 51.5 & 84.3 & 37.7 & 58.1 \\
\leftskip=1em\relax Label-routed MOPD & 55.3 & 63.2 & \textbf{56.8} & \textbf{52.6} & 84.0 & 38.4 & 58.4 \\
\rowcolor{qblue}\leftskip=1em\relax \textbf{DN-MOPD (Ours)} & \textbf{58.9} & \textbf{67.7} & 56.3 & 51.4 & \textbf{84.5} & \textbf{38.7} & \textbf{59.6} \\
\bottomrule
\end{tabularx}\endgroup\par
\par\vspace{4pt}\begingroup\fontsize{8}{9.6}\selectfont\raggedright
\textbf{Bold}: best within the multi-teacher OPD block.
\par\endgroup
\end{table}

\textbf{Experimental setup.} We use Qwen3.5-9B, 4B and 2B with
independently trained specialist pools at each size. Students learn from
teachers at their own size. The main OPD comparisons share experts,
student initialization and prompts within each pool, and use 80 updates,
student seed 42 (seeds 43 and 44 in \hyperref[app:C.5]{App. C.5}) and an 8,192-token
training response cap. The earlier Qwen3 results use different teacher
pools and protocols and are reported separately in \hyperref[app:E.2]{App. E.2}.

\textbf{Baselines and variants.} We report the initial student, all
three RL experts and OPD students trained with each single expert.
Uniform pooling averages teacher probabilities, Dynamic selects a
teacher from the current student response, and Label uses the prompt's
domain. DN-MOPD is compared with Label under the same OPD setup.
Annealed injection adds early imitation of teacher answers to Label
(Table \ref{tab:assignment_scale}, \hyperref[app:C.1]{App. C.1}). SeqKD-SFT
\citep{arxiv160607947} trains for 84 updates on fixed teacher answers
generated with a 16K cap, while parameter averaging and task arithmetic
\citep{arxiv221204089} combine expert weights directly; task arithmetic
uses \(\lambda=1\). These recipes differ in token exposure and compute
(\hyperref[app:A.2]{App. A.2}). The strongest single-teacher student is selected on the same
public Total, favoring that comparator.

\textbf{Evaluation.} AIME25/AIME26 measure mathematics, LiveCodeBench
(LCB) v5/v6 measure code generation on 167/175 disjoint problems, and
IFEval/IFBench measure instruction following. Task scores average
correctness first over sampled answers for each question, then over
questions, estimating single-answer accuracy. We sample 64, 6 and 16
responses per question for math, code and IF, respectively. IF uses
strict prompt accuracy. Domain scores and mean response lengths equally
average their two suite means; Total and the overall cap-hit rate
equally average all six suites. The main tables use a 16,384-token
evaluation cap, chosen after inspecting both caps; complete six-task
results at 8K and 16K are in \hyperref[app:B]{App. B}. Paired 95\% confidence intervals
resample questions within each suite and are conditional on the student
training seed and teacher pool. \hyperref[app:A]{App. A} gives training configurations and
scoring details.

\begin{table}[t]\centering
\caption{\textbf{Capability integration at smaller scales.} Independently trained Qwen3.5-4B and 2B expert pools. Each domain averages its two tasks; Total weights the three domains equally.}\label{tab:scales}
\definecolor{qgray}{RGB}{244,245,246}
\definecolor{qblue}{RGB}{233,242,249}
\begingroup\fontsize{9}{10.5}\selectfont\renewcommand{\arraystretch}{1.03}\setlength{\tabcolsep}{5.5pt}
\begin{tabularx}{\linewidth}{@{}>{\raggedright\arraybackslash}Xrrrrrrrr@{}}
\toprule
& \multicolumn{4}{c}{Qwen3.5-4B} & \multicolumn{4}{c}{Qwen3.5-2B} \\
\cmidrule(lr){2-5}\cmidrule(lr){6-9}
Method & Math & Code & IF & Total & Math & Code & IF & Total \\
\midrule
Initial student & 52.2 & 38.1 & 52.8 & 47.7 & 17.6 & 11.3 & 43.3 & 24.0 \\
\addlinespace[2pt]\rowcolor{qgray}\multicolumn{9}{@{}l@{}}{\textit{RL experts}} \\
\leftskip=1em\relax Math expert & 54.2 & 39.7 & 54.0 & 49.3 & 22.1 & 13.4 & 44.7 & 26.7 \\
\leftskip=1em\relax Code expert & 51.6 & 47.4 & 53.6 & 50.8 & 20.7 & 21.6 & 44.4 & 28.9 \\
\leftskip=1em\relax IF expert & 50.7 & 38.1 & 60.6 & 49.8 & 15.6 & 12.5 & 52.8 & 27.0 \\
\addlinespace[2pt]\rowcolor{qgray}\multicolumn{9}{@{}l@{}}{\textit{Offline distillation}} \\
\leftskip=1em\relax SeqKD-SFT & 55.6 & 52.3 & 59.7 & 55.9 & 23.1 & 23.9 & 51.6 & 32.9 \\
\addlinespace[2pt]\rowcolor{qgray}\multicolumn{9}{@{}l@{}}{\textit{Parameter merging}} \\
\leftskip=1em\relax ParamMerge-Avg & 54.5 & 43.1 & 56.2 & 51.2 & 20.7 & 16.3 & 47.5 & 28.2 \\
\leftskip=1em\relax ParamMerge-TA & 53.8 & 44.4 & 61.4 & 53.2 & 24.4 & 20.4 & 53.3 & 32.7 \\
\addlinespace[2pt]\rowcolor{qgray}\multicolumn{9}{@{}l@{}}{\textit{Single-teacher OPD}} \\
\leftskip=1em\relax Math teacher & 53.9 & 39.9 & 53.6 & 49.1 & 22.1 & 14.1 & 44.2 & 26.8 \\
\leftskip=1em\relax Code teacher & 52.2 & 46.8 & 53.5 & 50.9 & 20.4 & 21.2 & 44.9 & 28.8 \\
\leftskip=1em\relax IF teacher & 50.3 & 39.2 & 57.6 & 49.0 & 16.5 & 11.7 & 49.2 & 25.8 \\
\addlinespace[2pt]\rowcolor{qgray}\multicolumn{9}{@{}l@{}}{\textit{Multi-teacher OPD}} \\
\leftskip=1em\relax Uniform pool & 52.3 & 41.5 & 55.0 & 49.6 & 19.9 & 14.6 & 46.2 & 26.9 \\
\leftskip=1em\relax Dynamic router & 49.9 & 44.7 & 57.1 & 50.6 & 17.3 & 13.8 & 48.5 & 26.5 \\
\leftskip=1em\relax Label-routed MOPD & 50.2 & 43.3 & 57.4 & 50.3 & 17.0 & 13.4 & 49.5 & 26.6 \\
\rowcolor{qblue}\leftskip=1em\relax \textbf{DN-MOPD (Ours)} & \textbf{54.0} & \textbf{45.4} & \textbf{58.2} & \textbf{52.5} & \textbf{20.7} & \textbf{16.3} & \textbf{49.9} & \textbf{29.0} \\
\bottomrule
\end{tabularx}\endgroup\par
\par\vspace{4pt}\begingroup\fontsize{8}{9.6}\selectfont\raggedright
\textbf{Bold}: best within the multi-teacher OPD block at each size.
\par\endgroup
\end{table}

\phantomsection\label{rq:1}\label{sec:3.1}

\subsection{Main results}\label{main-results}

DN-MOPD improves on Label at every model size and both evaluation
budgets (\hyperref[fig:1]{Figure 1}, \hyperref[app:B]{App. B}), with paired intervals consistently above
zero (\hyperref[app:C.3]{App. C.3}). This comparison holds the experts, student
initialization, prompts and update budget fixed, isolating the change in
feedback weighting. The same normalization rule transfers across
independently trained expert pools without size-specific tuning. Across
three student seeds, every seed-matched comparison favors DN-MOPD (mean
gains +1.12, +1.97 and +2.34 at 16K, intervals above zero; \hyperref[app:C.5]{App. C.5}).

The main gain is in mathematics. At 16K, Label shows no mathematics gain
over the initial student at any size, whereas DN-MOPD improves
mathematics at every size (\hyperref[app:E.1]{App. E.1}). On MATH-500, Label likewise does
not exceed the initial student at 16K, while DN-MOPD leads Label by
+0.74, +1.25 and +5.95 points (\hyperref[app:C.7]{App. C.7}). Matching prompts to the right
specialist therefore leaves room to improve how its feedback is used.
Gains in code and instruction following are smaller and less consistent;
at 9B, code does not improve at 16K.

The single-teacher students provide a stronger reference than the
initial model: learning from one specialist can already yield a
competitive student across domains. Multi-teacher integration should
therefore be evaluated against this alternative as well as Label. Label
never exceeds the strongest single-teacher student, whereas DN-MOPD
exceeds it at every size. This lead is smaller than the gain over Label,
with some intervals including zero; at 9B it grows to +2.57 {[}+1.65,
+3.46{]} when both continue to 160 updates (\hyperref[app:C.3]{App. C.3}). The clearest
evidence is thus for improving the label-routed recipe; SeqKD-SFT and
task arithmetic retain higher Totals under their own training recipes
(\hyperref[tab:main]{Tables 1}--\hyperref[tab:scales]{2}).

\begin{figure}[t]\centering\includegraphics[width=.9\linewidth]{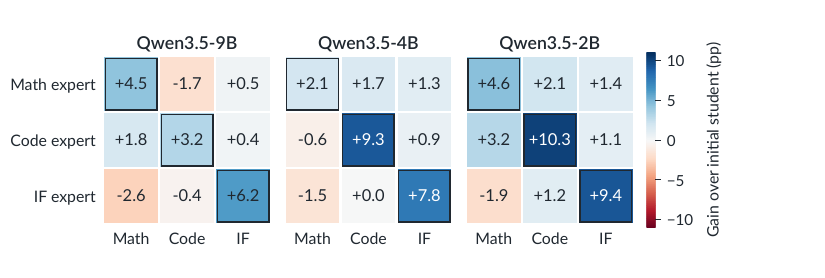}\caption{Expert specialization across domains. Each cell is the gain over the shared initialization (pp) at 16K, averaged over the two tasks in the evaluation domain. Rows identify experts and columns identify evaluation domains.}\label{fig:3}\end{figure}

\phantomsection\label{rq:2}\label{sec:3.2}

\subsection{Understanding the
improvement}\label{understanding-the-improvement}

\hyperref[fig:3]{Figure 3} establishes that the teachers have specialist capabilities
available to transfer: each expert improves its own domain, while its
effects on other domains vary. Yet Label fails to retain the mathematics
gain. Mathematics-only OPD retains much more of that gain (\hyperref[tab:main]{Tables 1}--\hyperref[tab:scales]{2}),
showing that the student can learn from this teacher in isolation. The
difficulty appears when its feedback is combined with feedback from the
other domains. \hyperref[app:E.1]{App. E.1} gives the detailed capability-transfer
comparison.

We next test whether changing teacher assignment can alleviate this
integration gap. Pooling combines teacher distributions, while dynamic
routing selects a teacher from the student's current response. Neither
brings consistent gains across sizes (\hyperref[tab:assignment_scale]{Table 3}). DN-MOPD instead retains
Label's assignment and changes the strength of each domain's feedback,
improving every size under the same teachers and training budget. This
comparison supports treating feedback scale as a separate design choice
from teacher selection. The two fixed-weight rows separate the
directions of DN-MOPD's adjustment at 4B and 2B. The intuitive remedy,
amplifying the under-transferred mathematics feedback while leaving IF
unchanged, recovers about half of DN-MOPD's improvement at 4B and little
at 2B. Reducing IF alone recovers most of it and raises mathematics by
about three points (\hyperref[app:C.6]{App. C.6}).

\begin{table}[t]\centering
\definecolor{qgray}{RGB}{244,245,246}
\definecolor{qblue}{RGB}{233,242,249}
\caption{\textbf{Changes to label-routed OPD.} Total point differences (pp) relative to Label, evaluated at 16K. Pooling and Dynamic change the teacher assignment; annealed injection and DN-MOPD retain it and change the learning signal. The two fixed-weight rows set one domain's weight (mathematics 2 or IF 0.25, others 1) and were run at 4B and 2B only.}\label{tab:assignment_scale}
\begingroup\fontsize{9}{11}\selectfont\setlength{\tabcolsep}{5pt}\renewcommand{\arraystretch}{1.1}%
\begin{tabular}{@{}lrrrrr@{}}
\toprule
Variant & Teacher rule & Signal change & 9B & 4B & 2B \\
\midrule
Uniform pool & Mixture & Unchanged & -0.41 & -0.69 & +0.30 \\
Dynamic router & Response & Unchanged & -0.25 & +0.28 & -0.07 \\
Label-routed & Domain & Unchanged & +0.00 & +0.00 & +0.00 \\
Annealed injection & Domain & Early imitation & -0.24 & +1.11 & +0.30 \\
Math $\times$2 only & Domain & Math scale & --- & +1.20 & +0.27 \\
IF $\times$0.25 only & Domain & IF scale & --- & +1.79 & +2.19 \\
\rowcolor{qblue}DN-MOPD & Domain & Domain scale & +1.17 & +2.24 & +2.36 \\
\bottomrule\end{tabular}\endgroup%
\par
\end{table}

Fixed weights for all three domains show where the benefit comes from
(\hyperref[app:C.6]{App. C.6}). Weights fixed at DN-MOPD's first-batch multipliers, or at a
global (2, 1, 0.25) chosen after inspecting them, show no detectable
difference from DN-MOPD at 16K at 9B and 4B; at 2B, per-batch estimation
outperforms the first-batch weights by 1.10 points. The gain therefore
comes mainly from calibrating the feedback scale, which DN-MOPD obtains
from batch statistics rather than from a per-size weight search.

\phantomsection\label{rq:3}

\begin{figure}[t]\centering\includegraphics[width=0.8\linewidth]{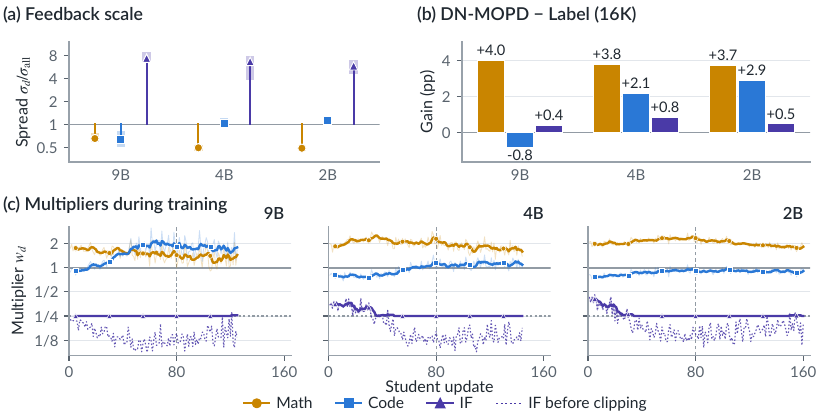}\caption{Feedback scales, domain gains and training multipliers. (a) Median relative log-ratio spread with interquartile ranges over the recorded batches of the DN-MOPD runs, including their continuation beyond 80 updates. (b) Domain accuracy differences between DN-MOPD and Label at 16K; point estimates. (c) Raw multipliers (faint), trailing 5-update means (solid), and IF before clipping (dotted). Vertical dashed lines mark update 80; traces end at updates 125, 144 and 160 for 9B, 4B and 2B.}\label{fig:4}\end{figure}

\hyperref[fig:4]{Figure 4} shows why this calibration matters. IF log-ratios are more
dispersed than the pooled signal and mathematics log-ratios less so
(panel a), and DN-MOPD amplifies mathematics and downweights IF
throughout training rather than only near initialization (panel c),
consistent with fixed first-batch weights working at the larger sizes.
IF supplies about 1\% of response tokens but up to half of the pooled
variance. For the initial 4B student it accounts for 94\% of the
combined gradient under equal weights and 64\% under DN-MOPD's
first-batch weights, and still 88\% if the loss is averaged over tokens
instead of responses (\hyperref[app:D.5]{App. D.5}). DN-MOPD thus appears to help mainly by
limiting this domain's influence on the shared update. After 80 updates
IF dominates the gradient under either weighting, but the DN-MOPD
student's mathematics and code log-ratios are about three times less
dispersed than Label's, consistent with more of those teachers' feedback
having been absorbed.

The mathematics gains come with shorter answers and fewer responses
reaching the generation cap (Table \ref{tab:lengths}), so they do not
come from longer generations; \hyperref[app:C.4]{App. C.4} reports both caps.

\begin{table}[t]\centering
\definecolor{qgray}{RGB}{244,245,246}
\definecolor{qblue}{RGB}{233,242,249}
\caption{\textbf{Capability and generated length.} Mean response tokens by domain and the share of responses reaching the 16K cap. Scores and lengths come from the same evaluation outputs.}\label{tab:lengths}
\begingroup\fontsize{9}{11}\selectfont\setlength{\tabcolsep}{5pt}\renewcommand{\arraystretch}{1.1}%
\begin{tabular}{@{}lrrrrr@{}}
\toprule
Method & Total & Math tokens & Code tokens & IF tokens & At cap (\%) \\
\midrule
\addlinespace[3pt]\rowcolor{qgray}\multicolumn{6}{@{}l@{}}{\textit{Qwen3.5-9B}} \\
\hspace*{1em}SeqKD-SFT & 62.6 & 5,547 & 5,896 & 422 & 1.5 \\
\hspace*{1em}ParamMerge-TA & 60.5 & 5,083 & 5,479 & 430 & 1.7 \\
\hspace*{1em}Label-routed & 58.4 & 8,637 & 6,846 & 450 & 9.8 \\
\rowcolor{qblue}\hspace*{1em}DN-MOPD & 59.6 & 7,032 & 6,427 & 441 & 5.2 \\
\addlinespace[3pt]\rowcolor{qgray}\multicolumn{6}{@{}l@{}}{\textit{Qwen3.5-4B}} \\
\hspace*{1em}SeqKD-SFT & 55.9 & 5,534 & 7,389 & 440 & 3.2 \\
\hspace*{1em}ParamMerge-TA & 53.2 & 4,865 & 6,332 & 494 & 2.9 \\
\hspace*{1em}Label-routed & 50.3 & 8,653 & 8,238 & 523 & 12.4 \\
\rowcolor{qblue}\hspace*{1em}DN-MOPD & 52.5 & 6,146 & 7,649 & 522 & 5.3 \\
\addlinespace[3pt]\rowcolor{qgray}\multicolumn{6}{@{}l@{}}{\textit{Qwen3.5-2B}} \\
\hspace*{1em}SeqKD-SFT & 32.9 & 6,670 & 9,419 & 822 & 8.9 \\
\hspace*{1em}ParamMerge-TA & 32.7 & 5,383 & 8,315 & 930 & 8.9 \\
\hspace*{1em}Label-routed & 26.6 & 10,246 & 10,047 & 744 & 20.2 \\
\rowcolor{qblue}\hspace*{1em}DN-MOPD & 29.0 & 7,444 & 10,054 & 706 & 13.5 \\
\bottomrule\end{tabular}\endgroup%
\par
\end{table}

\phantomsection\label{rq:4}\label{sec:3.3}

\subsection{Training budget and
scope}\label{training-budget-and-scope}

\begin{table}[t]\centering
\definecolor{qgray}{RGB}{244,245,246}
\definecolor{qblue}{RGB}{233,242,249}
\caption{\textbf{Training-duration comparison.} Six-task Total at fixed 80- and 160-update endpoints, evaluated at 16K. Changes use unrounded scores.}\label{tab:budget}
\begingroup\fontsize{9}{11}\selectfont\setlength{\tabcolsep}{5pt}\renewcommand{\arraystretch}{1.1}%
\begin{tabular}{@{}lrrr@{}}
\toprule
Method & 80 & 160 & $\Delta$ Total \\
\midrule
\addlinespace[3pt]\rowcolor{qgray}\multicolumn{4}{@{}l@{}}{\textit{Qwen3.5-9B}} \\
\hspace*{1em}Single teacher (IF) & 57.3 & 57.6 & +0.4 \\
\hspace*{1em}Label-routed & 58.4 & 59.4 & +1.0 \\
\rowcolor{qblue}\hspace*{1em}DN-MOPD & 59.6 & 60.7 & +1.2 \\
\addlinespace[3pt]\rowcolor{qgray}\multicolumn{4}{@{}l@{}}{\textit{Qwen3.5-4B}} \\
\hspace*{1em}Single teacher (code) & 50.9 & 51.2 & +0.4 \\
\hspace*{1em}Label-routed & 50.3 & 51.6 & +1.3 \\
\rowcolor{qblue}\hspace*{1em}DN-MOPD & 52.5 & 53.1 & +0.5 \\
\addlinespace[3pt]\rowcolor{qgray}\multicolumn{4}{@{}l@{}}{\textit{Qwen3.5-2B}} \\
\hspace*{1em}Single teacher (code) & 28.8 & 28.8 & -0.1 \\
\hspace*{1em}Label-routed & 26.6 & 28.3 & +1.7 \\
\rowcolor{qblue}\hspace*{1em}DN-MOPD & 29.0 & 30.0 & +1.1 \\
\bottomrule\end{tabular}\endgroup%
\par
\end{table}

We vary the evaluation and training budgets to check whether the
advantage depends on limited generation room or an early training
endpoint. Allowing longer evaluation responses helps Label more,
narrowing the gap, but DN-MOPD remains ahead across sizes with positive
paired intervals (\hyperref[app:C.3]{App. C.3}). Its advantage survives the larger
generation budget, although the margin depends on the allowed response
length.

The training extension tests whether Label catches up when given more
updates (Table \ref{tab:budget}). Each run continues its own checkpoint
with the same recipe; the single-teacher comparators were selected on
the development instrument (IF at 9B, code at 4B and 2B). The longer
runs narrow the gaps at the two smaller sizes in Table \ref{tab:budget},
but DN-MOPD remains ahead at every size under both evaluation caps. Its
advantage therefore persists beyond the main training endpoint, although
the eventual ordering at convergence remains unresolved.

\section{Related work}\label{sec:4}\label{related-work}

\textbf{On-Policy Distillation.} Supervising student-generated responses
addresses the training--inference mismatch of fixed teacher-written
corpora. MiniLLM \citep{arxiv230608543} uses reverse KL, generalized
knowledge distillation \citep{arxiv230613649} supports alternative
divergences, and DistiLLM \citep{arxiv240203898} combines skew-KL
targets with adaptive response reuse. Later methods refine the feedback:
Uni-OPD \citep{arxiv260503677} uses outcome calibration and improved
exploration; CROP \citep{arxiv260813387} selects task-relevant
supervision positions. Lightning OPD 2.0 \citep{arxiv260728449} removes
a recurring style component from teacher--reference disagreement, while
PowerOPD \citep{arxiv260617199} transforms sampled-token rewards to
control their magnitude. ExOPD \citep{arxiv260212125} changes the
balance between reward and KL regularization, studying reward
extrapolation in both single- and multi-teacher settings. DN-MOPD
rescales the ordinary sampled-token log-ratio by domain after teacher
assignment, leaving teacher probabilities unchanged.

\textbf{Multi-Teacher On-Policy Distillation.} MOPD
\citep{arxiv260630406} integrates independently trained domain
specialists into a shared student. Open-MOPD \citep{arxiv260819098}
examines capability imbalance under fixed domain assignments, addressing
token allocation, domain budgets and reward refresh. Multi-teacher OPD
also appears in specialized applications: UI-MOPD \citep{arxiv260704425}
integrates platform-specific GUI teachers, and LS-MOPD
\citep{arxiv260803610} integrates language specialists for multilingual
speech recognition. Our focus is the relative strength of domain
feedback under a fixed assignment and prompt schedule. DN-MOPD rescales
distillation advantages by the spread of each teacher's log-ratios
rather than by token share or mean reward magnitude, complementing
decisions about which expert teaches and how much weight each domain's
loss receives.

\textbf{Balancing learning signals across objectives and domains.}
Multi-task learning weights losses by gradient norms
\citep{arxiv171102257} or learned uncertainty \citep{arxiv170507115},
PopArt normalizes value targets across reward scales
\citep{arxiv160207714}, and Dr.~GRPO \citep{arxiv250320783} and DAPO
\citep{arxiv250314476} revisit how RL advantages and losses are
normalized over responses and tokens. DTO-KD \citep{hayder2026dtokd}
balances task and distillation losses at the gradient level, and DRPO
\citep{arxiv250600711} scales rewards by domain rarity and difficulty.
DN-MOPD addresses imbalance across teachers: it rescales distillation
advantages by the measured spread of each teacher's log-ratios on
current student answers, without learned weights or per-domain
gradients, while retaining the loss, teacher assignment and data mix.

\section{Conclusion and limitations}\label{sec:5}\label{conclusion-and-limitations}

MOPD integrates RL-trained experts by routing each prompt to the expert
of its domain, but in our Qwen3.5 constructions routing alone did not
make the student better than its strongest single-teacher counterpart,
and mathematics barely transferred. DN-MOPD adds the missing control: it
rescales each domain's distillation advantages by the batch-wise spread
of its teacher--student log-ratios. Across Qwen3.5-9B, 4B and 2B and
three student seeds, it improves the six-task average over MOPD at both
evaluation caps and recovers most of the lost mathematics gain. Controls
attribute most of this gain to limiting the dispersed
instruction-following feedback, which otherwise dominates the shared
gradient of the initial student. Integrating RL experts thus requires
calibrating how much each expert's feedback counts.

\textbf{Scope and limitations.} The comparisons use one expert pool per
size within one model family; the three student seeds do not cover
retraining the experts. DN-MOPD was selected on an earlier Qwen3
development instrument (\hyperref[app:A.4]{App. A.4}), and an earlier Qwen3-4B comparison
under a different setup found no clear gain (\hyperref[app:E.2]{App. E.2}), so benefits
depend on the teacher--student configuration. The scale estimates depend
on domain composition and response length, and the clipping bounds were
not tuned per size; the instruction-following multiplier usually sits at
the lower bound, so the rule bounds rather than equalizes that domain's
scale. Fixed weights near DN-MOPD's measured multipliers perform
comparably at 9B and 4B, so per-batch re-estimation adds to the gain
only at 2B.

\section*{AI use statement}

We used large language model (LLM) agents in this work, under our
direction, in three roles. For the experiments, they wrote the training
launchers, job supervisors, evaluation and analysis scripts, launched
and monitored the runs. For the writing, they drafted, revised and
shortened the text and formatted tables and figures. For checking, they
searched the literature, verified each reference against its arXiv or
publisher record, re-derived the reported numbers from the stored result
files, and reviewed experiment protocols and analysis code
adversarially. Every output was treated as a draft: reported values,
chart marks and confidence intervals are computed by scripts from stored
model outputs, and we read and approved the final text. We chose the
research questions, approved each experimental design and decision rule
before it ran, decided which experiments to continue or stop and what to
report, and drew the conclusions; we are responsible for everything that
appears here.

\section*{Reproducibility statement}

The paper and its appendices document each stage in the order it runs.
\hyperref[sec:2]{Section 2}, Algorithm 1 and \hyperref[app:D.1]{App. D.1} define DN-MOPD, including the
statistic conventions and where the multiplier enters the loss. \hyperref[app:A.1]{App. A.1}
gives the models, training data, and expert and student training
configurations; \hyperref[app:A.2]{App. A.2} specifies the baselines; \hyperref[app:A.3]{App. A.3} gives the
question lists, sampling settings, generation caps, grading and
bootstrap procedure behind every reported score; and \hyperref[app:A.4]{App. A.4} states how
the method was selected. Apps. B and C report the complete results and
controls, and \hyperref[app:D]{App. D} defines the feedback and gradient diagnostics.

Our code, training data and evaluation records are released at
\url{https://github.com/LiXin97/DN-MOPD}. The release contains the per-question correctness of all
80 evaluated Qwen3.5 models on the six tasks at both caps, with hashes
that link each score to its model, question manifest and grader
revision, together with the MATH-500 records and the training-time
multiplier traces. It includes a script that recomputes every Qwen3.5
score, domain mean, Total and paired interval from these records,
checking each score against its stored summary and each reported
contrast and interval against the values in the paper, a standalone
reference implementation of DN-MOPD with unit tests, and the
training code with recipes for every row of the tables. Scores can therefore be
recomputed without rerunning any model. Teacher and student checkpoints
will be released later.

\section*{Ethics statement}

The work distils public open-weight models on mathematics, code and
instruction-following data; it involves no human subjects and no
personal data. The code judge executes model-generated programs; it was
run in a resource-limited sandbox on completions generated by our own
rollouts only.

\bibliographystyle{iclr2027_conference}
\bibliography{refs}
\newpage
\clearpage
\appendix
\section*{Appendix overview}\label{appendix-overview}
\addcontentsline{toc}{section}{Appendix overview}

The appendices provide the experimental details and
additional evidence for DN-MOPD. \hyperref[app:A]{Appendix A} specifies the
training recipes, baselines, evaluation and statistical
protocol. \hyperref[app:B]{Appendix B} expands the main results into complete
task-level tables at both evaluation budgets. \hyperref[app:C]{Appendix C}
reports the annealed-injection and longer-training
comparisons, generation-length and truncation diagnostics,
additional student seeds, fixed-domain-weight controls and
MATH-500. \hyperref[app:D]{Appendix D} describes the exact scaling operation
and the measurements underlying the feedback analysis,
including domain token shares, clipping frequency, the
sources of the pooled spread and per-domain gradients.
\hyperref[app:E]{Appendix E} quantifies mathematics capability transfer and an
earlier construction in which normalization yields no clear
improvement.

\makeatletter\setlength{\@fptop}{0pt}\setlength{\@fpsep}{12pt}\makeatother
\vspace{0.6em}
\noindent\textbf{Contents}\par\medskip
\begingroup\renewcommand{\arraystretch}{1.5}
\begin{tabularx}{\linewidth}{@{}lXr@{}}
 & & Page \\
\hyperref[app:A]{A} & \hyperref[app:A]{Experimental setup and reproducibility} & \pageref{app:A} \\
\hyperref[app:B]{B} & \hyperref[app:B]{Complete six-task results} & \pageref{app:B} \\
\hyperref[app:C]{C} & \hyperref[app:C]{Additional controls and training budgets} & \pageref{app:C} \\
\hyperref[app:D]{D} & \hyperref[app:D]{Implementation and diagnostic details} & \pageref{app:D} \\
\hyperref[app:E]{E} & \hyperref[app:E]{Capability transfer and a boundary case} & \pageref{app:E} \\
\end{tabularx}\endgroup

\clearpage

\section{Experimental setup and
reproducibility}\label{app:A}\label{appendix-a-experimental-setup-and-reproducibility}

\setcounter{table}{0}\renewcommand{\thetable}{A.\arabic{table}}\renewcommand{\theHtable}{A.\arabic{table}}

\textbf{\phantomsection\label{app:A.1}A.1 Models, data and training.} We construct
separate pools of mathematics, code and
instruction-following specialists from Qwen3.5-9B, 4B and 2B
\citep{qwen35}. Each student starts from the corresponding
initial model and learns from specialists at its own size.
The teacher data draw mathematics prompts from DeepMath-103K
\citep{arxiv250411456}, code prompts from Eurus
\citep{arxiv240402078}, and generated instruction-following
prompts. Student training uses a fixed mixture of 2,700
prompts, 900 per domain, with the domain label identifying
the corresponding teacher.

The experts use GRPO \citep{arxiv240203300} with temperature
1.0, a 2,048-token prompt limit and an 8,192-token response
cap. Candidate rollout batches contain 128 prompts with
eight responses per prompt; the optimizer global batch is
256 responses. Dynamic sampling filters prompt groups
without reward variation, with at most eight generation
batches per rollout. Adam uses learning rate \(10^{-6}\),
constant after ten warm-up updates, betas \((0.9,0.98)\),
weight decay 0.1 and gradient clipping at 1.0. Teacher
training uses seed 42. Expert training ran for up to 400
updates or until a fixed wall-clock budget for the
construction ended, and we use the last complete checkpoint;
update counts therefore differ across pools (Table \hyperref[app:A.1]{A.1}).
Each roster is shared by every student comparison at that
size.

\begin{table}[!htbp]\centering
\definecolor{qgray}{RGB}{244,245,246}
\definecolor{qblue}{RGB}{233,242,249}
\caption{\textbf{Shared specialist checkpoints.} Update counts for the domain experts used by every student comparison within each size. Each specialist is trained from the same initialization as its student.}\label{tab:teacher:roster}
\begingroup\fontsize{9}{11}\selectfont\setlength{\tabcolsep}{5pt}\renewcommand{\arraystretch}{1.1}
\begin{tabular}{@{}lrrr@{}}\toprule
Size & Math expert & Code expert & IF expert \\\midrule
9B & 250 & 200 & 400 \\
4B & 320 & 300 & 400 \\
2B & 400 & 400 & 400 \\
\bottomrule\end{tabular}\endgroup
\par\end{table}

All main OPD students use 80 updates and training seed 42.
Each rollout batch contains 64 prompts with eight responses
per prompt, giving a global batch of 512 responses. The
prompt and response limits are 2,048 and 8,192 tokens,
respectively; rollout temperature is 1.0. Adam uses a
learning rate of \(10^{-6}\), held constant after five
warm-up updates, betas \((0.9,0.98)\), weight decay 0.1 and
gradient clipping at 1.0. The learning signal is the
sampled-token reverse-KL advantage inside the clipped
policy-gradient surrogate. \hyperref[app:D]{App. D} specifies where DN-MOPD
changes that signal.

\textbf{\phantomsection\label{app:A.2}A.2 Baselines.} Single-teacher OPD uses one
specialist for every prompt. Label selects the specialist
matching the prompt's domain. Uniform pooling averages
teacher probabilities, while Dynamic selects the teacher
with the lowest mean log-probability on the current student
response. These comparisons share the student
initialization, expert roster, prompt mixture and OPD
training budget. The strongest single-teacher reference in
the main results is selected on the same public Total,
favoring that comparator. The continuation study instead
uses its development-selected teacher (IF at 9B; code at 4B
and 2B).

SeqKD-SFT trains on fixed answers from the domain-matched
teachers, generated with a 16,384-token response cap, for 84
updates. This corresponds to approximately four passes over
one fixed answer per training prompt, using a global batch
of 128, learning rate \(10^{-5}\) with cosine decay to
\(10^{-6}\), and five warm-up updates. The optimizer uses
weight decay 0.1, betas \((0.9,0.98)\) and gradient clipping
at 1.0. The teacher answer corpus is reused across updates,
whereas OPD generates fresh student responses. Parameter
averaging \citep{arxiv220305482} combines the expert weights
directly. Task arithmetic adds the sum of
expert-minus-initialization parameter differences to the
initial model with coefficient 1.0. SFT, merging and OPD
therefore use different token and optimization budgets;
their scores compare capability under the stated recipes.
Annealed injection and 160-update continuations are
specified in \hyperref[app:C]{App. C}.

\textbf{\phantomsection\label{app:A.3}A.3 Evaluation and uncertainty.} AIME25 and AIME26
each contain 30 questions with 64 sampled answers per
question. LiveCodeBench v5/v6 \citep{arxiv240307974} contain
167/175 disjoint problems with six answers each. IFEval-541
\citep{arxiv231107911} and IFBench-300
\citep{arxiv250702833} use 16 answers per prompt and strict
prompt-level correctness. The fixed question lists, prompt
rendering, answer extraction, grader revisions and sample
identities accompany the score records. Evaluation uses the
non-thinking prompt rendering, temperature and top-p 1.0,
and generation seed 42. The same checkpoints are evaluated
at response caps of 8,192 and 16,384 tokens. Mathematics
answers are graded with Math-Verify on the last boxed
expression, and a missing boxed answer counts as incorrect.
Code is run against the official LiveCodeBench tests with
its official runner; each program runs in a separate process
with an 8 GiB memory limit and a 6-second limit per test,
and a response is correct only if it passes every test.
IFEval and IFBench responses are scored with the benchmarks'
official strict instruction checkers.

Task scores average correctness over answers within each
question and then over questions. Thus avg@N estimates
single-answer accuracy, rather than pass@N. Each domain
score equally averages its two suites; Total equally
averages all six. Response lengths are averaged within each
suite before taking domain means, and the overall cap-hit
rate equally averages the six suite rates. Displayed scores
are percentages; differences use unrounded values and are
expressed in percentage points.

Paired 95\% intervals use 10,000 bootstrap replicates with
seed 20260923, resampling questions within each suite while
preserving method pairing. Repeated answers to one question
are not treated as independent questions. These intervals
are conditional on one student training seed and one expert
pool per size; they do not quantify variation from
retraining students or experts.

\textbf{\phantomsection\label{app:A.4}A.4 Method selection and reporting.} The
domain-scaling operation and its clipping bounds were
selected on an earlier Qwen3 development instrument at the
80-update port-selection point, then transferred unchanged
to all three Qwen3.5 sizes. The later 160-update and
annealed comparisons are separate experiments. The earlier
Qwen3 public comparison yielded no clear gain over Label and
is retained in \hyperref[app:E]{App. E}.

The original Qwen3.5 public protocol specified an 8K
evaluation cap. The choice to show 16K in the main text was
made after both caps had been inspected, to match the
generation budget of the earlier Qwen3 public comparison.
This was a presentation amendment, not a preregistered
preference for 16K. The original internal development
comparisons kept their own 8K evaluation protocol; their
scores are not substituted for public-suite cells. \hyperref[app:B]{App. B}
retains complete 8K results for all methods and tasks. The
feedback diagnostics in \hyperref[app:D]{App. D} are descriptive analyses of
recorded training signals.

\textbf{\phantomsection\label{app:A.5}A.5 Reproduction materials.} The released records
(\url{https://github.com/LiXin97/DN-MOPD}) associate each score with its model, checkpoint,
training configuration, question manifest and grader
revision. Task scores, domain means, paired contrasts and
plotting data are derived from stored model outputs.
Machine-readable task scores and paired contrasts accompany
the tables.

\clearpage

\section{Complete six-task
results}\label{app:B}\label{appendix-b-complete-six-task-results}

The following tables expand the domain summaries in the main
text. The 9B results at 16K are already reported in \hyperref[tab:main]{Table 1};
\hyperref[app:B.1]{B.1} supplies the corresponding 4B and 2B tables. \hyperref[app:B.2]{B.2} reports
all three sizes at the originally specified 8K evaluation
cap. Every method uses the same checkpoint at both
evaluation caps. Scores are percentages, and boldface
identifies the best result within the multi-teacher OPD
block.

\setcounter{table}{0}\renewcommand{\thetable}{B.\arabic{table}}\renewcommand{\theHtable}{B.\arabic{table}}

\textbf{\phantomsection\label{app:B.1}B.1 Full six-task results for the smaller scales at
16K.} At 4B and 2B, DN-MOPD has the highest Total in the
multi-teacher OPD block and the best score on every task,
tying Dynamic on LCB v6 at 4B. Relative to Label, AIME25 and
AIME26 improve by 2.9 to 4.5 points and both LCB versions by
1.7 to 3.1 points, while IFEval and IFBench change by at
most 1.2 points. SeqKD-SFT and task arithmetic retain higher
Totals at both sizes, as in \hyperref[tab:scales]{Table 2}.

\begin{table}[!htbp]\centering
\definecolor{qgray}{RGB}{244,245,246}
\definecolor{qblue}{RGB}{233,242,249}
\caption{\textbf{Complete results for Qwen3.5-4B at 16K.} Scores (\%) on the six public tasks; all OPD students use 80 updates. Total equally averages the tasks. Boldface identifies the best result within the multi-teacher OPD block.}\label{tab:q35:4b:16384}
\begingroup\fontsize{9}{10.5}\selectfont\renewcommand{\arraystretch}{1.03}\setlength{\tabcolsep}{3pt}
\begin{tabularx}{\linewidth}{@{}>{\raggedright\arraybackslash}Xrrrrrrr@{}}
\toprule
& \multicolumn{2}{c}{Math (avg@64)} & \multicolumn{2}{c}{Code (avg@6)} & \multicolumn{2}{c}{IF (avg@16)} & \\
\cmidrule(lr){2-3}\cmidrule(lr){4-5}\cmidrule(lr){6-7}
Method & AIME25 & AIME26 & LCB v5 & LCB v6 & IFEval & IFBench & Total \\
\midrule
Initial student & 48.4 & 56.0 & 39.1 & 37.0 & 78.3 & 27.2 & 47.7 \\
\addlinespace[2pt]\rowcolor{qgray}\multicolumn{8}{@{}l@{}}{\textit{RL experts}} \\
\leftskip=1em\relax Math expert & 50.6 & 57.9 & 41.1 & 38.4 & 79.0 & 29.0 & 49.3 \\
\leftskip=1em\relax Code expert & 48.5 & 54.6 & 49.5 & 45.2 & 79.2 & 28.0 & 50.8 \\
\leftskip=1em\relax IF expert & 47.9 & 53.4 & 38.4 & 37.8 & 84.3 & 36.9 & 49.8 \\
\addlinespace[2pt]\rowcolor{qgray}\multicolumn{8}{@{}l@{}}{\textit{Offline distillation}} \\
\leftskip=1em\relax SeqKD-SFT & 51.9 & 59.3 & 55.2 & 49.4 & 83.2 & 36.2 & 55.9 \\
\addlinespace[2pt]\rowcolor{qgray}\multicolumn{8}{@{}l@{}}{\textit{Parameter merging}} \\
\leftskip=1em\relax ParamMerge-Avg & 50.3 & 58.6 & 46.7 & 39.4 & 81.0 & 31.3 & 51.2 \\
\leftskip=1em\relax ParamMerge-TA & 49.9 & 57.8 & 45.3 & 43.4 & 84.4 & 38.3 & 53.2 \\
\addlinespace[2pt]\rowcolor{qgray}\multicolumn{8}{@{}l@{}}{\textit{Single-teacher OPD}} \\
\leftskip=1em\relax Math teacher & 49.9 & 57.8 & 42.4 & 37.4 & 78.7 & 28.5 & 49.1 \\
\leftskip=1em\relax Code teacher & 47.7 & 56.6 & 49.5 & 44.2 & 78.4 & 28.7 & 50.9 \\
\leftskip=1em\relax IF teacher & 46.6 & 54.0 & 40.1 & 38.2 & 81.8 & 33.4 & 49.0 \\
\addlinespace[2pt]\rowcolor{qgray}\multicolumn{8}{@{}l@{}}{\textit{Multi-teacher OPD}} \\
\leftskip=1em\relax Uniform pool & 48.3 & 56.3 & 43.6 & 39.4 & 80.0 & 30.0 & 49.6 \\
\leftskip=1em\relax Dynamic router & 46.0 & 53.9 & 46.6 & \textbf{42.9} & 81.4 & 32.7 & 50.6 \\
\leftskip=1em\relax Label-routed MOPD & 46.0 & 54.4 & 46.3 & 40.3 & 81.6 & 33.2 & 50.3 \\
\rowcolor{qblue}\leftskip=1em\relax \textbf{DN-MOPD (Ours)} & \textbf{50.2} & \textbf{57.8} & \textbf{48.0} & \textbf{42.9} & \textbf{82.0} & \textbf{34.4} & \textbf{52.5} \\
\bottomrule
\end{tabularx}\endgroup\par
\par\end{table}

\begin{table}[!htbp]\centering
\definecolor{qgray}{RGB}{244,245,246}
\definecolor{qblue}{RGB}{233,242,249}
\caption{\textbf{Complete results for Qwen3.5-2B at 16K.} Scores (\%) on the six public tasks; all OPD students use 80 updates. Total equally averages the tasks. Boldface identifies the best result within the multi-teacher OPD block.}\label{tab:q35:2b:16384}
\begingroup\fontsize{9}{10.5}\selectfont\renewcommand{\arraystretch}{1.03}\setlength{\tabcolsep}{3pt}
\begin{tabularx}{\linewidth}{@{}>{\raggedright\arraybackslash}Xrrrrrrr@{}}
\toprule
& \multicolumn{2}{c}{Math (avg@64)} & \multicolumn{2}{c}{Code (avg@6)} & \multicolumn{2}{c}{IF (avg@16)} & \\
\cmidrule(lr){2-3}\cmidrule(lr){4-5}\cmidrule(lr){6-7}
Method & AIME25 & AIME26 & LCB v5 & LCB v6 & IFEval & IFBench & Total \\
\midrule
Initial student & 17.7 & 17.4 & 9.0 & 13.5 & 62.8 & 23.9 & 24.0 \\
\addlinespace[2pt]\rowcolor{qgray}\multicolumn{8}{@{}l@{}}{\textit{RL experts}} \\
\leftskip=1em\relax Math expert & 20.9 & 23.3 & 10.1 & 16.7 & 64.1 & 25.3 & 26.7 \\
\leftskip=1em\relax Code expert & 19.6 & 21.8 & 20.1 & 23.0 & 63.5 & 25.3 & 28.9 \\
\leftskip=1em\relax IF expert & 16.0 & 15.2 & 10.0 & 15.0 & 74.6 & 30.9 & 27.0 \\
\addlinespace[2pt]\rowcolor{qgray}\multicolumn{8}{@{}l@{}}{\textit{Offline distillation}} \\
\leftskip=1em\relax SeqKD-SFT & 21.0 & 25.3 & 21.7 & 26.2 & 71.9 & 31.3 & 32.9 \\
\addlinespace[2pt]\rowcolor{qgray}\multicolumn{8}{@{}l@{}}{\textit{Parameter merging}} \\
\leftskip=1em\relax ParamMerge-Avg & 19.4 & 22.0 & 13.8 & 18.9 & 68.7 & 26.2 & 28.2 \\
\leftskip=1em\relax ParamMerge-TA & 22.7 & 26.1 & 18.6 & 22.3 & 74.7 & 31.9 & 32.7 \\
\addlinespace[2pt]\rowcolor{qgray}\multicolumn{8}{@{}l@{}}{\textit{Single-teacher OPD}} \\
\leftskip=1em\relax Math teacher & 20.8 & 23.3 & 11.6 & 16.7 & 63.7 & 24.6 & 26.8 \\
\leftskip=1em\relax Code teacher & 19.9 & 20.9 & 18.1 & 24.3 & 64.8 & 25.0 & 28.8 \\
\leftskip=1em\relax IF teacher & 16.4 & 16.6 & 10.4 & 13.0 & 70.0 & 28.4 & 25.8 \\
\addlinespace[2pt]\rowcolor{qgray}\multicolumn{8}{@{}l@{}}{\textit{Multi-teacher OPD}} \\
\leftskip=1em\relax Uniform pool & 19.2 & 20.7 & 11.8 & 17.4 & 66.9 & 25.5 & 26.9 \\
\leftskip=1em\relax Dynamic router & 16.9 & 17.8 & 11.3 & 16.3 & 69.1 & 27.8 & 26.5 \\
\leftskip=1em\relax Label-routed MOPD & 16.8 & 17.1 & 11.4 & 15.4 & 70.0 & 28.9 & 26.6 \\
\rowcolor{qblue}\leftskip=1em\relax \textbf{DN-MOPD (Ours)} & \textbf{19.7} & \textbf{21.6} & \textbf{14.5} & \textbf{18.1} & \textbf{70.2} & \textbf{29.6} & \textbf{29.0} \\
\bottomrule
\end{tabularx}\endgroup\par
\par\end{table}

\textbf{\phantomsection\label{app:B.2}B.2 Complete results at the declared 8K evaluation
cap.} The ordering is unchanged at the shorter budget.
DN-MOPD has the highest Total in the multi-teacher OPD block
at every size and the best score on five or six of the six
tasks; Label is marginally ahead on IFEval at 9B and 2B, by
0.06 and 0.10 points. The AIME gains over Label are larger
than at 16K, from 3.6 to 9.0 points (computed from unrounded
scores), consistent with the larger Total advantage at 8K
(\hyperref[app:C.3]{App. C.3}). SeqKD-SFT and task arithmetic again have higher
Totals.

\begin{table}[!htbp]\centering
\definecolor{qgray}{RGB}{244,245,246}
\definecolor{qblue}{RGB}{233,242,249}
\caption{\textbf{Complete results for Qwen3.5-9B at 8K.} Scores (\%) on the six public tasks; all OPD students use 80 updates. Total equally averages the tasks. Boldface identifies the best result within the multi-teacher OPD block.}\label{tab:q35:9b:8192}
\begingroup\fontsize{9}{10.5}\selectfont\renewcommand{\arraystretch}{1.03}\setlength{\tabcolsep}{3pt}
\begin{tabularx}{\linewidth}{@{}>{\raggedright\arraybackslash}Xrrrrrrr@{}}
\toprule
& \multicolumn{2}{c}{Math (avg@64)} & \multicolumn{2}{c}{Code (avg@6)} & \multicolumn{2}{c}{IF (avg@16)} & \\
\cmidrule(lr){2-3}\cmidrule(lr){4-5}\cmidrule(lr){6-7}
Method & AIME25 & AIME26 & LCB v5 & LCB v6 & IFEval & IFBench & Total \\
\midrule
Initial student & 45.5 & 50.7 & 45.0 & 42.2 & 82.0 & 34.0 & 49.9 \\
\addlinespace[2pt]\rowcolor{qgray}\multicolumn{8}{@{}l@{}}{\textit{RL experts}} \\
\leftskip=1em\relax Math expert & 56.2 & 64.6 & 45.5 & 44.3 & 82.5 & 35.0 & 54.7 \\
\leftskip=1em\relax Code expert & 48.3 & 55.7 & 52.1 & 46.4 & 82.2 & 34.3 & 53.2 \\
\leftskip=1em\relax IF expert & 45.2 & 50.0 & 45.1 & 43.9 & 86.4 & 41.9 & 52.1 \\
\addlinespace[2pt]\rowcolor{qgray}\multicolumn{8}{@{}l@{}}{\textit{Offline distillation}} \\
\leftskip=1em\relax SeqKD-SFT & 58.7 & 66.7 & 55.8 & 49.9 & 85.9 & 42.6 & 59.9 \\
\addlinespace[2pt]\rowcolor{qgray}\multicolumn{8}{@{}l@{}}{\textit{Parameter merging}} \\
\leftskip=1em\relax ParamMerge-Avg & 50.1 & 60.5 & 48.0 & 45.2 & 83.9 & 36.1 & 54.0 \\
\leftskip=1em\relax ParamMerge-TA & 55.9 & 66.4 & 50.4 & 47.6 & 86.2 & 43.5 & 58.3 \\
\addlinespace[2pt]\rowcolor{qgray}\multicolumn{8}{@{}l@{}}{\textit{Single-teacher OPD}} \\
\leftskip=1em\relax Math teacher & 54.6 & 63.0 & 48.1 & 44.6 & 81.9 & 35.0 & 54.5 \\
\leftskip=1em\relax Code teacher & 49.6 & 57.0 & 50.6 & 46.5 & 82.5 & 34.1 & 53.4 \\
\leftskip=1em\relax IF teacher & 45.6 & 48.9 & 44.7 & 41.7 & 84.6 & 38.8 & 50.7 \\
\addlinespace[2pt]\rowcolor{qgray}\multicolumn{8}{@{}l@{}}{\textit{Multi-teacher OPD}} \\
\leftskip=1em\relax Uniform pool & 49.2 & 55.8 & 47.7 & 44.2 & 83.2 & 36.0 & 52.7 \\
\leftskip=1em\relax Dynamic router & 46.6 & 53.1 & 47.7 & 44.4 & 84.1 & 37.7 & 52.2 \\
\leftskip=1em\relax Label-routed MOPD & 47.7 & 54.6 & 47.5 & 44.6 & \textbf{84.3} & 38.3 & 52.8 \\
\rowcolor{qblue}\leftskip=1em\relax \textbf{DN-MOPD (Ours)} & \textbf{51.3} & \textbf{60.0} & \textbf{51.1} & \textbf{46.0} & 84.3 & \textbf{39.2} & \textbf{55.3} \\
\bottomrule
\end{tabularx}\endgroup\par
\par\end{table}

\begin{table}[!htbp]\centering
\definecolor{qgray}{RGB}{244,245,246}
\definecolor{qblue}{RGB}{233,242,249}
\caption{\textbf{Complete results for Qwen3.5-4B at 8K.} Scores (\%) on the six public tasks; all OPD students use 80 updates. Total equally averages the tasks. Boldface identifies the best result within the multi-teacher OPD block.}\label{tab:q35:4b:8192}
\begingroup\fontsize{9}{10.5}\selectfont\renewcommand{\arraystretch}{1.03}\setlength{\tabcolsep}{3pt}
\begin{tabularx}{\linewidth}{@{}>{\raggedright\arraybackslash}Xrrrrrrr@{}}
\toprule
& \multicolumn{2}{c}{Math (avg@64)} & \multicolumn{2}{c}{Code (avg@6)} & \multicolumn{2}{c}{IF (avg@16)} & \\
\cmidrule(lr){2-3}\cmidrule(lr){4-5}\cmidrule(lr){6-7}
Method & AIME25 & AIME26 & LCB v5 & LCB v6 & IFEval & IFBench & Total \\
\midrule
Initial student & 37.8 & 43.0 & 30.6 & 32.1 & 78.5 & 27.6 & 41.6 \\
\addlinespace[2pt]\rowcolor{qgray}\multicolumn{8}{@{}l@{}}{\textit{RL experts}} \\
\leftskip=1em\relax Math expert & 49.4 & 57.1 & 34.0 & 33.2 & 79.4 & 28.9 & 47.0 \\
\leftskip=1em\relax Code expert & 42.3 & 47.7 & 43.5 & 41.5 & 79.3 & 28.3 & 47.1 \\
\leftskip=1em\relax IF expert & 37.4 & 41.8 & 31.3 & 32.9 & 84.4 & 36.7 & 44.1 \\
\addlinespace[2pt]\rowcolor{qgray}\multicolumn{8}{@{}l@{}}{\textit{Offline distillation}} \\
\leftskip=1em\relax SeqKD-SFT & 51.0 & 57.9 & 48.2 & 43.6 & 83.2 & 36.0 & 53.3 \\
\addlinespace[2pt]\rowcolor{qgray}\multicolumn{8}{@{}l@{}}{\textit{Parameter merging}} \\
\leftskip=1em\relax ParamMerge-Avg & 44.4 & 52.9 & 37.5 & 34.0 & 81.3 & 31.8 & 47.0 \\
\leftskip=1em\relax ParamMerge-TA & 49.4 & 56.8 & 41.0 & 39.0 & 84.1 & 38.5 & 51.5 \\
\addlinespace[2pt]\rowcolor{qgray}\multicolumn{8}{@{}l@{}}{\textit{Single-teacher OPD}} \\
\leftskip=1em\relax Math teacher & 47.7 & 56.7 & 32.8 & 33.3 & 78.5 & 29.1 & 46.3 \\
\leftskip=1em\relax Code teacher & 42.1 & 48.9 & 43.1 & 38.0 & 78.5 & 28.6 & 46.5 \\
\leftskip=1em\relax IF teacher & 38.6 & 42.9 & 31.0 & 31.9 & 81.8 & 33.6 & 43.3 \\
\addlinespace[2pt]\rowcolor{qgray}\multicolumn{8}{@{}l@{}}{\textit{Multi-teacher OPD}} \\
\leftskip=1em\relax Uniform pool & 42.5 & 49.8 & 33.7 & 33.3 & 79.8 & 29.9 & 44.8 \\
\leftskip=1em\relax Dynamic router & 39.2 & 44.9 & 38.3 & 34.8 & 81.4 & 32.2 & 45.1 \\
\leftskip=1em\relax Label-routed MOPD & 40.5 & 44.0 & 37.8 & 35.1 & 81.8 & 33.6 & 45.5 \\
\rowcolor{qblue}\leftskip=1em\relax \textbf{DN-MOPD (Ours)} & \textbf{46.4} & \textbf{52.9} & \textbf{39.7} & \textbf{35.8} & \textbf{82.1} & \textbf{34.4} & \textbf{48.6} \\
\bottomrule
\end{tabularx}\endgroup\par
\par\end{table}

\begin{table}[!htbp]\centering
\definecolor{qgray}{RGB}{244,245,246}
\definecolor{qblue}{RGB}{233,242,249}
\caption{\textbf{Complete results for Qwen3.5-2B at 8K.} Scores (\%) on the six public tasks; all OPD students use 80 updates. Total equally averages the tasks. Boldface identifies the best result within the multi-teacher OPD block.}\label{tab:q35:2b:8192}
\begingroup\fontsize{9}{10.5}\selectfont\renewcommand{\arraystretch}{1.03}\setlength{\tabcolsep}{3pt}
\begin{tabularx}{\linewidth}{@{}>{\raggedright\arraybackslash}Xrrrrrrr@{}}
\toprule
& \multicolumn{2}{c}{Math (avg@64)} & \multicolumn{2}{c}{Code (avg@6)} & \multicolumn{2}{c}{IF (avg@16)} & \\
\cmidrule(lr){2-3}\cmidrule(lr){4-5}\cmidrule(lr){6-7}
Method & AIME25 & AIME26 & LCB v5 & LCB v6 & IFEval & IFBench & Total \\
\midrule
Initial student & 11.1 & 8.0 & 7.6 & 12.0 & 63.6 & 24.1 & 21.1 \\
\addlinespace[2pt]\rowcolor{qgray}\multicolumn{8}{@{}l@{}}{\textit{RL experts}} \\
\leftskip=1em\relax Math expert & 19.6 & 21.6 & 8.4 & 15.2 & 63.5 & 25.2 & 25.6 \\
\leftskip=1em\relax Code expert & 17.8 & 17.3 & 15.7 & 19.9 & 63.9 & 24.8 & 26.6 \\
\leftskip=1em\relax IF expert & 12.1 & 8.2 & 8.0 & 12.8 & 74.8 & 30.5 & 24.4 \\
\addlinespace[2pt]\rowcolor{qgray}\multicolumn{8}{@{}l@{}}{\textit{Offline distillation}} \\
\leftskip=1em\relax SeqKD-SFT & 20.6 & 22.0 & 21.9 & 25.1 & 71.7 & 31.4 & 32.1 \\
\addlinespace[2pt]\rowcolor{qgray}\multicolumn{8}{@{}l@{}}{\textit{Parameter merging}} \\
\leftskip=1em\relax ParamMerge-Avg & 16.4 & 16.0 & 10.5 & 16.9 & 69.1 & 26.3 & 25.8 \\
\leftskip=1em\relax ParamMerge-TA & 22.3 & 24.6 & 18.6 & 21.5 & 74.6 & 31.9 & 32.3 \\
\addlinespace[2pt]\rowcolor{qgray}\multicolumn{8}{@{}l@{}}{\textit{Single-teacher OPD}} \\
\leftskip=1em\relax Math teacher & 19.5 & 20.9 & 9.1 & 13.5 & 63.9 & 24.9 & 25.3 \\
\leftskip=1em\relax Code teacher & 16.6 & 17.8 & 17.2 & 21.6 & 64.3 & 25.1 & 27.1 \\
\leftskip=1em\relax IF teacher & 11.9 & 9.1 & 7.5 & 11.3 & 69.6 & 28.5 & 23.0 \\
\addlinespace[2pt]\rowcolor{qgray}\multicolumn{8}{@{}l@{}}{\textit{Multi-teacher OPD}} \\
\leftskip=1em\relax Uniform pool & 16.1 & 15.1 & 9.2 & 14.4 & 66.7 & 25.3 & 24.5 \\
\leftskip=1em\relax Dynamic router & 14.3 & 11.8 & 8.9 & 14.6 & 69.5 & 28.2 & 24.5 \\
\leftskip=1em\relax Label-routed MOPD & 14.5 & 11.7 & 9.9 & 14.4 & \textbf{70.3} & 29.0 & 25.0 \\
\rowcolor{qblue}\leftskip=1em\relax \textbf{DN-MOPD (Ours)} & \textbf{19.1} & \textbf{19.8} & \textbf{11.5} & \textbf{17.0} & 70.2 & \textbf{29.6} & \textbf{27.9} \\
\bottomrule
\end{tabularx}\endgroup\par
\par\end{table}

\clearpage

\section{Additional controls and training
budgets}\label{app:C}\label{appendix-c-additional-controls-and-training-budgets}

\setcounter{table}{0}\renewcommand{\thetable}{C.\arabic{table}}\renewcommand{\theHtable}{C.\arabic{table}}

\textbf{\phantomsection\label{app:C.1}C.1 Annealed injection.} This control augments Label
with imitation of verified-correct answers from the
domain-matched teacher. For each prompt covered by the fixed
answer bank, one of the eight response slots is replaced by
a teacher answer, prioritizing an incorrect student
response. Other slots retain the Label learning signal.
Injected tokens receive a constant advantage with weight
\(\max(0,1-t/40)\) at training step \(t\); injection stops
at step 40. The student is evaluated at 80 updates. Bank
coverage varies across pools, so the intervention applies to
covered prompts rather than every prompt. Table \hyperref[app:C.1]{C.1} reports
this coverage so that the strength of the intervention can
be interpreted across sizes. A covered prompt has at least
one usable verified teacher answer; coverage is not the
proportion of training tokens supplied by a teacher.

\begin{table}[!htbp]\centering
\definecolor{qgray}{RGB}{244,245,246}
\definecolor{qblue}{RGB}{233,242,249}
\caption{\textbf{Teacher-answer coverage for annealed injection.} Prompts with at least one usable verified-correct answer, out of the 900 training prompts in each domain. One response slot is replaced for covered groups while the imitation weight is positive; uncovered prompts retain Label training.}\label{tab:anneal:coverage}
\begingroup\fontsize{9}{11}\selectfont\setlength{\tabcolsep}{5pt}\renewcommand{\arraystretch}{1.1}
\begin{tabular}{@{}lrrr@{}}\toprule
Size & Math & Code & IF \\\midrule
9B & 792/900 & 724/900 & 812/900 \\
4B & 794/900 & 686/900 & 763/900 \\
2B & 599/900 & 363/900 & 771/900 \\
\bottomrule\end{tabular}\endgroup
\par\end{table}

The comparison tests early teacher imitation as an
alternative change to the supervision; fixed domain weights
are tested separately in \hyperref[app:C.6]{App. C.6}. At 16K, annealed
injection changes the Total by \(-0.24\), \(+1.11\) and
\(+0.30\) points relative to Label at 9B, 4B and 2B, so its
gains depend on model size, and it trails DN-MOPD at every
size (Table \ref{tab:assignment_scale}).

\textbf{\phantomsection\label{app:C.2}C.2 Longer training.} Label, DN-MOPD and a
single-teacher reference continue their own runs from 80 to
160 updates with the same recipe. The single teacher is
selected on the development instrument: IF at 9B, code at 4B
and 2B. The tables below give all six task scores at both
evaluation caps, including the completed continuations
behind Table \ref{tab:budget}. They test whether the
ordering persists beyond the main endpoint, without assuming
that either method has converged.

\begin{table}[!htbp]\centering
\definecolor{qgray}{RGB}{244,245,246}
\definecolor{qblue}{RGB}{233,242,249}
\caption{\textbf{Additional controls at 16K.} Scores (\%); each block includes three 160-update continuations and the 80-update annealed-injection control. All runs use seed 42. The single-teacher reference is selected on the development instrument.}\label{tab:extra:16384}
\begingroup\fontsize{9}{10.5}\selectfont\setlength{\tabcolsep}{3pt}\renewcommand{\arraystretch}{1.03}
\begin{tabularx}{\linewidth}{@{}>{\raggedright\arraybackslash}Xrrrrrrrr@{}}\toprule
Method & Updates & AIME25 & AIME26 & LCB v5 & LCB v6 & IFEval & IFBench & Total \\\midrule
\addlinespace[2pt]\rowcolor{qgray}\multicolumn{9}{@{}l@{}}{\textit{Qwen3.5-9B}} \\
\leftskip=1em\relax Label & 160 & 57.0 & 63.9 & 58.9 & 51.0 & 84.9 & 40.6 & 59.4 \\
\rowcolor{qblue}\leftskip=1em\relax DN-MOPD & 160 & 58.9 & 70.1 & 56.7 & 52.5 & 85.1 & 41.1 & 60.7 \\
\leftskip=1em\relax Single IF & 160 & 55.8 & 61.7 & 52.4 & 50.1 & 85.1 & 40.6 & 57.6 \\
\leftskip=1em\relax Annealed injection & 80 & 55.8 & 63.6 & 54.3 & 52.0 & 84.6 & 38.6 & 58.2 \\
\addlinespace[2pt]\rowcolor{qgray}\multicolumn{9}{@{}l@{}}{\textit{Qwen3.5-4B}} \\
\leftskip=1em\relax Label & 160 & 47.6 & 54.2 & 47.3 & 42.6 & 82.8 & 35.2 & 51.6 \\
\rowcolor{qblue}\leftskip=1em\relax DN-MOPD & 160 & 50.3 & 58.1 & 47.2 & 44.6 & 82.3 & 35.9 & 53.1 \\
\leftskip=1em\relax Single code & 160 & 48.3 & 55.9 & 51.0 & 44.5 & 78.9 & 28.6 & 51.2 \\
\leftskip=1em\relax Annealed injection & 80 & 47.6 & 55.5 & 47.0 & 43.2 & 81.9 & 33.3 & 51.4 \\
\addlinespace[2pt]\rowcolor{qgray}\multicolumn{9}{@{}l@{}}{\textit{Qwen3.5-2B}} \\
\leftskip=1em\relax Label & 160 & 16.4 & 19.1 & 13.2 & 18.8 & 71.3 & 31.0 & 28.3 \\
\rowcolor{qblue}\leftskip=1em\relax DN-MOPD & 160 & 20.0 & 23.1 & 14.5 & 20.6 & 71.5 & 30.6 & 30.0 \\
\leftskip=1em\relax Single code & 160 & 19.8 & 20.7 & 18.9 & 24.0 & 63.9 & 25.3 & 28.8 \\
\leftskip=1em\relax Annealed injection & 80 & 18.1 & 17.4 & 11.1 & 16.5 & 70.0 & 28.4 & 26.9 \\
\bottomrule\end{tabularx}\endgroup
\par\end{table}

\begin{table}[!htbp]\centering
\definecolor{qgray}{RGB}{244,245,246}
\definecolor{qblue}{RGB}{233,242,249}
\caption{\textbf{Additional controls at 8K.} Scores (\%); each block includes three 160-update continuations and the 80-update annealed-injection control. All runs use seed 42. The single-teacher reference is selected on the development instrument.}\label{tab:extra:8192}
\begingroup\fontsize{9}{10.5}\selectfont\setlength{\tabcolsep}{3pt}\renewcommand{\arraystretch}{1.03}
\begin{tabularx}{\linewidth}{@{}>{\raggedright\arraybackslash}Xrrrrrrrr@{}}\toprule
Method & Updates & AIME25 & AIME26 & LCB v5 & LCB v6 & IFEval & IFBench & Total \\\midrule
\addlinespace[2pt]\rowcolor{qgray}\multicolumn{9}{@{}l@{}}{\textit{Qwen3.5-9B}} \\
\leftskip=1em\relax Label & 160 & 47.3 & 54.5 & 48.6 & 44.5 & 85.1 & 40.3 & 53.4 \\
\rowcolor{qblue}\leftskip=1em\relax DN-MOPD & 160 & 52.2 & 62.3 & 50.3 & 45.8 & 85.1 & 41.2 & 56.2 \\
\leftskip=1em\relax Single IF & 160 & 45.9 & 51.3 & 43.7 & 43.2 & 85.3 & 41.0 & 51.8 \\
\leftskip=1em\relax Annealed injection & 80 & 46.4 & 52.0 & 45.1 & 42.9 & 84.5 & 38.2 & 51.5 \\
\addlinespace[2pt]\rowcolor{qgray}\multicolumn{9}{@{}l@{}}{\textit{Qwen3.5-4B}} \\
\leftskip=1em\relax Label & 160 & 40.7 & 47.3 & 38.3 & 34.8 & 82.6 & 36.0 & 46.6 \\
\rowcolor{qblue}\leftskip=1em\relax DN-MOPD & 160 & 46.0 & 54.9 & 42.9 & 37.7 & 82.6 & 36.0 & 50.0 \\
\leftskip=1em\relax Single code & 160 & 41.0 & 47.4 & 44.4 & 39.4 & 79.1 & 28.6 & 46.7 \\
\leftskip=1em\relax Annealed injection & 80 & 39.6 & 44.8 & 36.0 & 34.9 & 81.8 & 33.3 & 45.1 \\
\addlinespace[2pt]\rowcolor{qgray}\multicolumn{9}{@{}l@{}}{\textit{Qwen3.5-2B}} \\
\leftskip=1em\relax Label & 160 & 15.0 & 14.7 & 8.8 & 16.4 & 71.1 & 31.0 & 26.2 \\
\rowcolor{qblue}\leftskip=1em\relax DN-MOPD & 160 & 18.6 & 19.9 & 11.6 & 17.8 & 71.6 & 30.8 & 28.4 \\
\leftskip=1em\relax Single code & 160 & 17.1 & 17.2 & 16.7 & 21.8 & 63.5 & 25.0 & 26.9 \\
\leftskip=1em\relax Annealed injection & 80 & 13.8 & 12.1 & 9.0 & 15.0 & 70.0 & 28.9 & 24.8 \\
\bottomrule\end{tabularx}\endgroup
\par\end{table}

Label and DN-MOPD both improve from 80 to 160 updates at
every size and cap, and DN-MOPD remains the highest within
every block. At the 8K evaluation cap, Label after 160
updates still does not reach DN-MOPD after 80 updates: the
80-update DN-MOPD student is ahead by 1.93, 1.95 and 1.71
points at 9B, 4B and 2B, with every interval above zero. At
16K these differences are smaller, +0.18 to +0.94, and their
intervals include or touch zero. Annealed injection at 80
updates is close to Label at 9B and 2B, higher at 4B, and
below DN-MOPD at every size (\hyperref[app:C.3]{App. C.3}).

\textbf{\phantomsection\label{app:C.3}C.3 Paired contrasts.} The paired intervals below
accompany the aggregate comparisons in the main text. They
use the question-level bootstrap described in \hyperref[app:A.3]{App. A.3}. The
DN-MOPD-minus-Label intervals remain above zero at both
endpoints and both evaluation caps. Comparisons involving
annealed injection use the 80-update checkpoints.

\begin{table}[!htbp]\centering
\definecolor{qgray}{RGB}{244,245,246}
\definecolor{qblue}{RGB}{233,242,249}
\caption{\textbf{Paired Total differences at 16K.} Differences and intervals are in pp. Parentheses give student update counts. The 95\% intervals use 10,000 paired question-bootstrap replicates, conditional on seed 42 and the shared expert pool.}\label{tab:extra:contrasts:16384}
\begingroup\fontsize{9}{11}\selectfont\setlength{\tabcolsep}{5pt}\renewcommand{\arraystretch}{1.1}
\begin{tabular}{@{}lrr@{}}\toprule
Contrast (updates) & $\Delta$ Total & 95\% CI \\\midrule
\addlinespace[2pt]\rowcolor{qgray}\multicolumn{3}{@{}l@{}}{\textit{Qwen3.5-9B}} \\
\rowcolor{qblue}\hspace*{1em}DN $-$ Label (80) & $+1.17$ & $[+0.28, +2.03]$ \\
\rowcolor{qblue}\hspace*{1em}DN $-$ Label (160) & $+1.33$ & $[+0.42, +2.22]$ \\
\hspace*{1em}Annealed $-$ Label (80) & $-0.24$ & $[-0.98, +0.51]$ \\
\hspace*{1em}DN $-$ Annealed (80) & $+1.41$ & $[+0.48, +2.34]$ \\
\addlinespace[2pt]\rowcolor{qgray}\multicolumn{3}{@{}l@{}}{\textit{Qwen3.5-4B}} \\
\rowcolor{qblue}\hspace*{1em}DN $-$ Label (80) & $+2.24$ & $[+1.30, +3.20]$ \\
\rowcolor{qblue}\hspace*{1em}DN $-$ Label (160) & $+1.46$ & $[+0.53, +2.40]$ \\
\hspace*{1em}Annealed $-$ Label (80) & $+1.11$ & $[+0.28, +1.95]$ \\
\hspace*{1em}DN $-$ Annealed (80) & $+1.12$ & $[+0.20, +2.04]$ \\
\addlinespace[2pt]\rowcolor{qgray}\multicolumn{3}{@{}l@{}}{\textit{Qwen3.5-2B}} \\
\rowcolor{qblue}\hspace*{1em}DN $-$ Label (80) & $+2.36$ & $[+1.58, +3.17]$ \\
\rowcolor{qblue}\hspace*{1em}DN $-$ Label (160) & $+1.77$ & $[+0.97, +2.60]$ \\
\hspace*{1em}Annealed $-$ Label (80) & $+0.30$ & $[-0.27, +0.87]$ \\
\hspace*{1em}DN $-$ Annealed (80) & $+2.06$ & $[+1.26, +2.90]$ \\
\bottomrule\end{tabular}\endgroup
\par\end{table}

\begin{table}[!htbp]\centering
\definecolor{qgray}{RGB}{244,245,246}
\definecolor{qblue}{RGB}{233,242,249}
\caption{\textbf{Paired Total differences at 8K.} Differences and intervals are in pp. Parentheses give student update counts. The 95\% intervals use 10,000 paired question-bootstrap replicates, conditional on seed 42 and the shared expert pool.}\label{tab:extra:contrasts:8192}
\begingroup\fontsize{9}{11}\selectfont\setlength{\tabcolsep}{5pt}\renewcommand{\arraystretch}{1.1}
\begin{tabular}{@{}lrr@{}}\toprule
Contrast (updates) & $\Delta$ Total & 95\% CI \\\midrule
\addlinespace[2pt]\rowcolor{qgray}\multicolumn{3}{@{}l@{}}{\textit{Qwen3.5-9B}} \\
\rowcolor{qblue}\hspace*{1em}DN $-$ Label (80) & $+2.47$ & $[+1.65, +3.27]$ \\
\rowcolor{qblue}\hspace*{1em}DN $-$ Label (160) & $+2.79$ & $[+1.93, +3.65]$ \\
\hspace*{1em}Annealed $-$ Label (80) & $-1.33$ & $[-2.04, -0.65]$ \\
\hspace*{1em}DN $-$ Annealed (80) & $+3.80$ & $[+2.91, +4.69]$ \\
\addlinespace[2pt]\rowcolor{qgray}\multicolumn{3}{@{}l@{}}{\textit{Qwen3.5-4B}} \\
\rowcolor{qblue}\hspace*{1em}DN $-$ Label (80) & $+3.08$ & $[+2.12, +4.06]$ \\
\rowcolor{qblue}\hspace*{1em}DN $-$ Label (160) & $+3.42$ & $[+2.43, +4.43]$ \\
\hspace*{1em}Annealed $-$ Label (80) & $-0.41$ & $[-1.13, +0.34]$ \\
\hspace*{1em}DN $-$ Annealed (80) & $+3.49$ & $[+2.48, +4.51]$ \\
\addlinespace[2pt]\rowcolor{qgray}\multicolumn{3}{@{}l@{}}{\textit{Qwen3.5-2B}} \\
\rowcolor{qblue}\hspace*{1em}DN $-$ Label (80) & $+2.92$ & $[+2.02, +3.85]$ \\
\rowcolor{qblue}\hspace*{1em}DN $-$ Label (160) & $+2.23$ & $[+1.46, +3.05]$ \\
\hspace*{1em}Annealed $-$ Label (80) & $-0.17$ & $[-0.69, +0.34]$ \\
\hspace*{1em}DN $-$ Annealed (80) & $+3.09$ & $[+2.17, +4.06]$ \\
\bottomrule\end{tabular}\endgroup
\par\end{table}

Table \ref{tab:single:contrasts} compares both label-based
recipes with the strongest single-teacher student, chosen at
each size and cap on the same public Total. Label is never
ahead of that student: the difference includes zero at 9B
and 4B at 16K and is negative with an interval excluding
zero in the other four cells. DN-MOPD is ahead in every
cell; its lead is separable in four of six cells and
includes zero at 9B at 8K and at 2B at 16K.

\begin{table}[!htbp]\centering
\definecolor{qgray}{RGB}{244,245,246}
\definecolor{qblue}{RGB}{233,242,249}
\caption{\textbf{Label and DN-MOPD against the strongest single-teacher student.} Paired six-task Total differences (pp) with 95\% question-bootstrap intervals, 80 updates. The strongest single-teacher student is chosen separately at each size and cap on the same public Total, which favors that comparator.}\label{tab:single:contrasts}
\begingroup\fontsize{9}{11}\selectfont\setlength{\tabcolsep}{5pt}\renewcommand{\arraystretch}{1.1}
\begin{tabular}{@{}llrr@{}}\toprule
Size & Strongest single & Label $-$ single & DN-MOPD $-$ single \\\midrule
\addlinespace[2pt]\rowcolor{qgray}\multicolumn{4}{@{}l@{}}{\textit{16K evaluation}} \\
\hspace*{1em}9B & Code & $-0.09$ {\scriptsize $[-1.00, +0.83]$} & $+1.08$ {\scriptsize $[+0.20, +1.98]$} \\
\hspace*{1em}4B & Code & $-0.55$ {\scriptsize $[-1.58, +0.46]$} & $+1.69$ {\scriptsize $[+0.77, +2.61]$} \\
\hspace*{1em}2B & Code & $-2.23$ {\scriptsize $[-3.17, -1.31]$} & $+0.13$ {\scriptsize $[-0.68, +0.93]$} \\
\addlinespace[2pt]\rowcolor{qgray}\multicolumn{4}{@{}l@{}}{\textit{8K evaluation}} \\
\hspace*{1em}9B & Math & $-1.69$ {\scriptsize $[-2.84, -0.58]$} & $+0.78$ {\scriptsize $[-0.13, +1.68]$} \\
\hspace*{1em}4B & Code & $-1.05$ {\scriptsize $[-1.96, -0.14]$} & $+2.03$ {\scriptsize $[+1.09, +2.96]$} \\
\hspace*{1em}2B & Code & $-2.12$ {\scriptsize $[-3.08, -1.21]$} & $+0.79$ {\scriptsize $[+0.01, +1.60]$} \\
\bottomrule\end{tabular}\endgroup
\par\end{table}

The strongest 9B single-teacher student at the main 16K cap,
trained from the code expert, was also continued to 160
updates. It does not improve with the additional updates
(the difference from its 80-update checkpoint includes zero
at both caps), whereas DN-MOPD's lead over it widens from
+1.08 at 80 updates to +2.57 at 160 updates at 16K, and from
+1.92 to +2.97 at 8K, with every interval above zero. Label
after 160 updates is ahead of it at 16K and tied at 8K.

\begin{table}[!htbp]\centering
\definecolor{qgray}{RGB}{244,245,246}
\definecolor{qblue}{RGB}{233,242,249}
\caption{\textbf{Continuing the strongest 9B single-teacher student.} The code-teacher student has the highest six-task Total among the 9B single-teacher students at the main 16K cap (at 8K the mathematics-teacher student is higher, 54.53 versus 53.39, and was not continued). It is continued from 80 to 160 updates with the same recipe, alongside the Label and DN-MOPD continuations of Table~\ref{tab:budget}. Paired six-task Total differences (pp) with 95\% question-bootstrap intervals.}\label{tab:e8}
\begingroup\fontsize{9}{11}\selectfont\setlength{\tabcolsep}{5pt}\renewcommand{\arraystretch}{1.1}
\begin{tabular}{@{}lrrrr@{}}\toprule
Cap & DN-MOPD $-$ code (80) & DN-MOPD $-$ code (160) & Label $-$ code (160) & Code: 160 $-$ 80 \\\midrule
16K & $+1.08$ {\scriptsize $[+0.20, +1.98]$} & $+2.57$ {\scriptsize $[+1.65, +3.46]$} & $+1.24$ {\scriptsize $[+0.22, +2.25]$} & $-0.34$ {\scriptsize $[-1.18, +0.53]$} \\
8K & $+1.92$ {\scriptsize $[+1.10, +2.71]$} & $+2.97$ {\scriptsize $[+2.10, +3.83]$} & $+0.18$ {\scriptsize $[-0.71, +1.05]$} & $-0.19$ {\scriptsize $[-0.92, +0.54]$} \\
\bottomrule\end{tabular}\endgroup
\par\end{table}

The following table resolves DN-MOPD minus Label by domain.
Mathematics improves at every size, cap and endpoint, with
every interval above zero. Code and IF differences are
smaller and mixed. At 9B, code is lower at 16K (\(-0.82\),
with an interval including zero) but higher at 8K (+2.51,
with an interval above zero). The domain pattern supports
the mathematics-led gain described in the main text; it does
not establish a code cost at 9B.

\begin{table}[!htbp]\centering
\definecolor{qgray}{RGB}{244,245,246}
\definecolor{qblue}{RGB}{233,242,249}
\caption{\textbf{DN-MOPD minus Label by domain.} Paired differences in each domain mean (pp) with 95\% question-bootstrap intervals, at the main 80-update endpoint and after continuing both runs to 160 updates.}\label{tab:domain:contrasts}
\begingroup\fontsize{9}{11}\selectfont\setlength{\tabcolsep}{5pt}\renewcommand{\arraystretch}{1.1}
\begin{tabular}{@{}lrrrr@{}}\toprule
Size & Updates & Math & Code & IF \\\midrule
\addlinespace[2pt]\rowcolor{qgray}\multicolumn{5}{@{}l@{}}{\textit{16K evaluation}} \\
\rowcolor{qblue}\hspace*{1em}9B & 80 & $+3.98$ {\scriptsize $[+2.32, +5.68]$} & $-0.82$ {\scriptsize $[-2.78, +1.06]$} & $+0.36$ {\scriptsize $[-0.09, +0.82]$} \\
\hspace*{1em}9B & 160 & $+4.06$ {\scriptsize $[+2.21, +5.99]$} & $-0.38$ {\scriptsize $[-2.27, +1.51]$} & $+0.31$ {\scriptsize $[-0.11, +0.74]$} \\
\rowcolor{qblue}\hspace*{1em}4B & 80 & $+3.75$ {\scriptsize $[+1.95, +5.60]$} & $+2.13$ {\scriptsize $[+0.02, +4.28]$} & $+0.82$ {\scriptsize $[+0.29, +1.38]$} \\
\hspace*{1em}4B & 160 & $+3.33$ {\scriptsize $[+1.56, +5.21]$} & $+0.95$ {\scriptsize $[-1.16, +3.09]$} & $+0.10$ {\scriptsize $[-0.38, +0.59]$} \\
\rowcolor{qblue}\hspace*{1em}2B & 80 & $+3.72$ {\scriptsize $[+2.03, +5.52]$} & $+2.88$ {\scriptsize $[+1.31, +4.44]$} & $+0.47$ {\scriptsize $[-0.09, +1.03]$} \\
\hspace*{1em}2B & 160 & $+3.83$ {\scriptsize $[+2.11, +5.68]$} & $+1.55$ {\scriptsize $[+0.00, +3.10]$} & $-0.08$ {\scriptsize $[-0.67, +0.51]$} \\
\addlinespace[2pt]\rowcolor{qgray}\multicolumn{5}{@{}l@{}}{\textit{8K evaluation}} \\
\rowcolor{qblue}\hspace*{1em}9B & 80 & $+4.51$ {\scriptsize $[+2.84, +6.17]$} & $+2.51$ {\scriptsize $[+0.78, +4.22]$} & $+0.40$ {\scriptsize $[-0.06, +0.89]$} \\
\hspace*{1em}9B & 160 & $+6.35$ {\scriptsize $[+4.48, +8.33]$} & $+1.51$ {\scriptsize $[-0.14, +3.14]$} & $+0.49$ {\scriptsize $[+0.06, +0.92]$} \\
\rowcolor{qblue}\hspace*{1em}4B & 80 & $+7.42$ {\scriptsize $[+5.21, +9.74]$} & $+1.28$ {\scriptsize $[-0.46, +3.04]$} & $+0.55$ {\scriptsize $[+0.04, +1.06]$} \\
\hspace*{1em}4B & 160 & $+6.43$ {\scriptsize $[+4.22, +8.78]$} & $+3.77$ {\scriptsize $[+1.78, +5.82]$} & $+0.05$ {\scriptsize $[-0.43, +0.53]$} \\
\rowcolor{qblue}\hspace*{1em}2B & 80 & $+6.35$ {\scriptsize $[+4.11, +8.80]$} & $+2.13$ {\scriptsize $[+0.72, +3.54]$} & $+0.26$ {\scriptsize $[-0.29, +0.81]$} \\
\hspace*{1em}2B & 160 & $+4.45$ {\scriptsize $[+2.60, +6.46]$} & $+2.11$ {\scriptsize $[+0.82, +3.47]$} & $+0.12$ {\scriptsize $[-0.39, +0.64]$} \\
\bottomrule\end{tabular}\endgroup
\par\end{table}

\textbf{\phantomsection\label{app:C.4}C.4 Generation length and truncation.} The main text
reports length diagnostics at 16K. The complete paired
measurements below cover both evaluation budgets, using
fresh generations at each cap. Mathematics improvements
accompany shorter answers at every size under both caps.
This weakens the explanation that DN-MOPD succeeds simply by
spending more output tokens, while leaving open whether
shorter answers are a cause or a consequence of better
solutions. Average length alone does not measure inference
latency or training cost.

\begin{table}[!htbp]\centering
\definecolor{qgray}{RGB}{244,245,246}
\definecolor{qblue}{RGB}{233,242,249}
\caption{\textbf{Generation length and cap hits at both budgets.} Mean output tokens for the 80-update students, averaged equally over the two suites in each domain. The final column is the cap-hit percentage averaged equally over all six suites. Each budget uses its own generated responses.}\label{tab:lengths:appendix}
\begingroup\fontsize{9}{11}\selectfont\setlength{\tabcolsep}{5pt}\renewcommand{\arraystretch}{1.1}
\begin{tabular}{@{}lrrrr@{}}\toprule
Method & Math tokens & Code tokens & IF tokens & Cap hits (\%) \\\midrule
\addlinespace[2pt]\rowcolor{qgray}\multicolumn{5}{@{}l@{}}{\textit{Qwen3.5-9B (16K evaluation)}} \\
\hspace*{1em}Label & 8,637 & 6,846 & 450 & 9.8 \\
\rowcolor{qblue}\hspace*{1em}DN-MOPD & 7,032 & 6,427 & 441 & 5.2 \\
\addlinespace[2pt]\rowcolor{qgray}\multicolumn{5}{@{}l@{}}{\textit{Qwen3.5-4B (16K evaluation)}} \\
\hspace*{1em}Label & 8,653 & 8,238 & 523 & 12.4 \\
\rowcolor{qblue}\hspace*{1em}DN-MOPD & 6,146 & 7,649 & 522 & 5.3 \\
\addlinespace[2pt]\rowcolor{qgray}\multicolumn{5}{@{}l@{}}{\textit{Qwen3.5-2B (16K evaluation)}} \\
\hspace*{1em}Label & 10,246 & 10,047 & 744 & 20.2 \\
\rowcolor{qblue}\hspace*{1em}DN-MOPD & 7,444 & 10,054 & 706 & 13.5 \\
\addlinespace[2pt]\rowcolor{qgray}\multicolumn{5}{@{}l@{}}{\textit{Qwen3.5-9B (8K evaluation)}} \\
\hspace*{1em}Label & 6,071 & 4,886 & 408 & 31.4 \\
\rowcolor{qblue}\hspace*{1em}DN-MOPD & 5,712 & 4,811 & 387 & 25.7 \\
\addlinespace[2pt]\rowcolor{qgray}\multicolumn{5}{@{}l@{}}{\textit{Qwen3.5-4B (8K evaluation)}} \\
\hspace*{1em}Label & 6,318 & 5,421 & 450 & 36.2 \\
\rowcolor{qblue}\hspace*{1em}DN-MOPD & 5,642 & 5,363 & 417 & 26.4 \\
\addlinespace[2pt]\rowcolor{qgray}\multicolumn{5}{@{}l@{}}{\textit{Qwen3.5-2B (8K evaluation)}} \\
\hspace*{1em}Label & 7,443 & 5,834 & 563 & 45.4 \\
\rowcolor{qblue}\hspace*{1em}DN-MOPD & 6,700 & 6,003 & 569 & 34.5 \\
\bottomrule\end{tabular}\endgroup
\par\end{table}

\textbf{\phantomsection\label{app:C.5}C.5 Additional student seeds.} The main comparison
uses student seed 42. Two further seed-matched pairs of
Label and DN-MOPD runs (seeds 43 and 44) change the prompt
order and rollout sampling while keeping the experts,
initialization, prompts and 80-update recipe fixed; the
seed-43 and seed-44 Label runs use the same training code as
DN-MOPD with every multiplier set to one. DN-MOPD is ahead
in all nine seed-matched comparisons at each cap, with every
per-seed interval above zero. The three-seed mean
differences at 16K are +1.12, +1.97 and +2.34 points at 9B,
4B and 2B, and their two-level intervals exclude zero. A
second seed-42 Label run, trained with the same code as the
seed-43 and seed-44 Label runs, differs from the original
seed-42 Label run by −0.32, +0.62 and +0.19 points in Total
at 16K, with every interval including zero.

\begin{table}[!htbp]\centering
\definecolor{qgray}{RGB}{244,245,246}
\definecolor{qblue}{RGB}{233,242,249}
\caption{\textbf{DN-MOPD minus Label across student seeds.} Six-task Total differences (pp) between seed-matched runs at 80 updates; each seed changes the prompt order and rollout sampling while the teachers and initialization are shared. Per-seed intervals resample questions within suite; the three-seed mean uses a two-level bootstrap over seeds and questions. Seed 42 is the main comparison in \hyperref[tab:main]{Tables 1}--\hyperref[tab:scales]{2}.}\label{tab:seeds}
\begingroup\fontsize{9}{11}\selectfont\setlength{\tabcolsep}{5pt}\renewcommand{\arraystretch}{1.1}
\begin{tabular}{@{}lrrrr@{}}\toprule
Size & Seed 42 & Seed 43 & Seed 44 & Three-seed mean \\\midrule
\addlinespace[2pt]\rowcolor{qgray}\multicolumn{5}{@{}l@{}}{\textit{16K evaluation}} \\
\hspace*{1em}9B & $+1.17$ {\scriptsize $[+0.28, +2.03]$} & $+1.08$ {\scriptsize $[+0.21, +1.96]$} & $+1.10$ {\scriptsize $[+0.23, +1.97]$} & $+1.12$ {\scriptsize $[+0.62, +1.62]$} \\
\hspace*{1em}4B & $+2.24$ {\scriptsize $[+1.30, +3.20]$} & $+1.85$ {\scriptsize $[+0.91, +2.84]$} & $+1.82$ {\scriptsize $[+0.88, +2.77]$} & $+1.97$ {\scriptsize $[+1.39, +2.56]$} \\
\hspace*{1em}2B & $+2.36$ {\scriptsize $[+1.58, +3.17]$} & $+2.24$ {\scriptsize $[+1.45, +3.04]$} & $+2.43$ {\scriptsize $[+1.53, +3.40]$} & $+2.34$ {\scriptsize $[+1.86, +2.84]$} \\
\addlinespace[2pt]\rowcolor{qgray}\multicolumn{5}{@{}l@{}}{\textit{8K evaluation}} \\
\hspace*{1em}9B & $+2.47$ {\scriptsize $[+1.65, +3.27]$} & $+2.62$ {\scriptsize $[+1.80, +3.44]$} & $+3.05$ {\scriptsize $[+2.14, +3.96]$} & $+2.71$ {\scriptsize $[+2.16, +3.29]$} \\
\hspace*{1em}4B & $+3.08$ {\scriptsize $[+2.12, +4.06]$} & $+2.74$ {\scriptsize $[+1.79, +3.72]$} & $+3.01$ {\scriptsize $[+2.06, +3.97]$} & $+2.94$ {\scriptsize $[+2.37, +3.53]$} \\
\hspace*{1em}2B & $+2.92$ {\scriptsize $[+2.02, +3.85]$} & $+2.88$ {\scriptsize $[+2.03, +3.77]$} & $+2.83$ {\scriptsize $[+1.92, +3.80]$} & $+2.87$ {\scriptsize $[+2.36, +3.40]$} \\
\bottomrule\end{tabular}\endgroup
\par\end{table}

\textbf{\phantomsection\label{app:C.6}C.6 Fixed domain weights.} These controls replace
DN-MOPD's per-batch multiplier with constants that are fixed
for the whole run, keeping Label assignment and the rest of
the recipe. The update-0 control uses the multipliers
DN-MOPD measured on its first batch at the same size, before
training could affect the student; the global control
applies one setting, (2, 1, 0.25), at every size. Two
single-component controls at 4B and 2B change only one
domain: mathematics amplified to 2 with IF at 1, or IF
reduced to 0.25 with mathematics at 1.

Every fixed-weight control that changes both domains
improves on Label at every size and cap. The update-0
weights show no detectable difference from DN-MOPD at 9B and
4B (intervals include zero) but trail it at 2B (\(-1.10\) at
16K and \(-1.42\) at 8K, intervals below zero): a
calibration measured once before training recovers the gain
at the two larger sizes, while per-batch estimation adds to
it at 2B. The global setting, chosen after inspecting
DN-MOPD's measured multipliers, shows no detectable
difference from DN-MOPD at 9B, ahead of it at 4B at 8K
(+0.85, interval above zero) and behind it at 2B at 8K
(\(-0.64\), interval below zero). Among the single-component
controls, reducing IF alone improves on Label at both sizes
and caps and raises the mathematics score by 2.76 and 3.07
points at 16K at 4B and 2B. Amplifying mathematics alone
helps less: it is ahead of Label at 4B at 16K and
indistinguishable from Label at 2B. Neither component alone
exceeds DN-MOPD. These controls use one seed and are read
against the seed-42 runs.

\begin{table}[!htbp]\centering
\definecolor{qgray}{RGB}{244,245,246}
\definecolor{qblue}{RGB}{233,242,249}
\caption{\textbf{Fixed domain weights.} Each control trains with Label assignment and a constant multiplier per domain (math / code / IF), with the DN-MOPD recipe otherwise unchanged (80 updates, seed 42). Update 0 uses the multipliers DN-MOPD measured on its first batch at that size; Global uses one setting at every size. Paired differences (pp) with 95\% question-bootstrap intervals; the last column is the mathematics domain mean.}\label{tab:fixed:weights}
\begingroup\fontsize{9}{11}\selectfont\setlength{\tabcolsep}{4pt}\renewcommand{\arraystretch}{1.1}
\begin{tabular}{@{}llrrrr@{}}\toprule
Size & Control & Weights & $-$ Label & $-$ DN-MOPD & Math $-$ Label \\\midrule
\addlinespace[2pt]\rowcolor{qgray}\multicolumn{6}{@{}l@{}}{\textit{16K evaluation}} \\
\hspace*{1em}9B & Update 0 & 1.77/0.89/0.25 & $+1.29$ {\scriptsize $[+0.40, +2.16]$} & $+0.12$ {\scriptsize $[-0.69, +0.93]$} & $+3.59$ {\scriptsize $[+1.77, +5.42]$} \\
\hspace*{1em}9B & Global & 2/1/0.25 & $+1.32$ {\scriptsize $[+0.36, +2.29]$} & $+0.14$ {\scriptsize $[-0.67, +0.95]$} & $+4.01$ {\scriptsize $[+2.01, +6.15]$} \\
\hspace*{1em}4B & Update 0 & 2.01/0.81/0.34 & $+2.34$ {\scriptsize $[+1.47, +3.24]$} & $+0.10$ {\scriptsize $[-0.62, +0.82]$} & $+3.96$ {\scriptsize $[+2.21, +5.83]$} \\
\hspace*{1em}4B & Global & 2/1/0.25 & $+2.86$ {\scriptsize $[+1.90, +3.84]$} & $+0.62$ {\scriptsize $[-0.09, +1.36]$} & $+4.53$ {\scriptsize $[+2.63, +6.51]$} \\
\hspace*{1em}4B & Math only & 2/1/1 & $+1.20$ {\scriptsize $[+0.34, +2.08]$} & $-1.04$ {\scriptsize $[-1.95, -0.14]$} & $+1.72$ {\scriptsize $[+0.08, +3.44]$} \\
\hspace*{1em}4B & IF only & 1/1/0.25 & $+1.79$ {\scriptsize $[+0.86, +2.76]$} & $-0.44$ {\scriptsize $[-1.35, +0.48]$} & $+2.76$ {\scriptsize $[+0.99, +4.58]$} \\
\hspace*{1em}2B & Update 0 & 2.10/0.82/0.43 & $+1.26$ {\scriptsize $[+0.52, +2.01]$} & $-1.10$ {\scriptsize $[-1.74, -0.47]$} & $+3.41$ {\scriptsize $[+1.72, +5.18]$} \\
\hspace*{1em}2B & Global & 2/1/0.25 & $+2.27$ {\scriptsize $[+1.45, +3.13]$} & $-0.09$ {\scriptsize $[-0.73, +0.58]$} & $+3.62$ {\scriptsize $[+1.72, +5.70]$} \\
\hspace*{1em}2B & Math only & 2/1/1 & $+0.27$ {\scriptsize $[-0.34, +0.88]$} & $-2.08$ {\scriptsize $[-2.86, -1.34]$} & $+1.02$ {\scriptsize $[-0.26, +2.29]$} \\
\hspace*{1em}2B & IF only & 1/1/0.25 & $+2.19$ {\scriptsize $[+1.37, +3.03]$} & $-0.17$ {\scriptsize $[-0.84, +0.52]$} & $+3.07$ {\scriptsize $[+1.35, +4.84]$} \\
\addlinespace[2pt]\rowcolor{qgray}\multicolumn{6}{@{}l@{}}{\textit{8K evaluation}} \\
\hspace*{1em}9B & Update 0 & 1.77/0.89/0.25 & $+1.92$ {\scriptsize $[+1.06, +2.78]$} & $-0.55$ {\scriptsize $[-1.28, +0.19]$} & $+4.77$ {\scriptsize $[+2.94, +6.64]$} \\
\hspace*{1em}9B & Global & 2/1/0.25 & $+2.39$ {\scriptsize $[+1.49, +3.33]$} & $-0.08$ {\scriptsize $[-0.82, +0.67]$} & $+4.92$ {\scriptsize $[+2.86, +7.01]$} \\
\hspace*{1em}4B & Update 0 & 2.01/0.81/0.34 & $+2.98$ {\scriptsize $[+2.09, +3.93]$} & $-0.10$ {\scriptsize $[-0.82, +0.62]$} & $+6.61$ {\scriptsize $[+4.61, +8.88]$} \\
\hspace*{1em}4B & Global & 2/1/0.25 & $+3.93$ {\scriptsize $[+2.95, +4.94]$} & $+0.85$ {\scriptsize $[+0.21, +1.50]$} & $+7.94$ {\scriptsize $[+5.63, +10.39]$} \\
\hspace*{1em}4B & Math only & 2/1/1 & $+0.65$ {\scriptsize $[-0.11, +1.43]$} & $-2.43$ {\scriptsize $[-3.37, -1.51]$} & $+1.17$ {\scriptsize $[-0.34, +2.76]$} \\
\hspace*{1em}4B & IF only & 1/1/0.25 & $+2.22$ {\scriptsize $[+1.29, +3.17]$} & $-0.86$ {\scriptsize $[-1.64, -0.09]$} & $+5.03$ {\scriptsize $[+3.10, +6.98]$} \\
\hspace*{1em}2B & Update 0 & 2.10/0.82/0.43 & $+1.49$ {\scriptsize $[+0.76, +2.25]$} & $-1.42$ {\scriptsize $[-2.05, -0.83]$} & $+3.85$ {\scriptsize $[+2.16, +5.70]$} \\
\hspace*{1em}2B & Global & 2/1/0.25 & $+2.28$ {\scriptsize $[+1.47, +3.12]$} & $-0.64$ {\scriptsize $[-1.14, -0.12]$} & $+4.84$ {\scriptsize $[+2.86, +6.95]$} \\
\hspace*{1em}2B & Math only & 2/1/1 & $+0.04$ {\scriptsize $[-0.50, +0.59]$} & $-2.88$ {\scriptsize $[-3.73, -2.04]$} & $+0.44$ {\scriptsize $[-0.44, +1.33]$} \\
\hspace*{1em}2B & IF only & 1/1/0.25 & $+1.73$ {\scriptsize $[+0.97, +2.52]$} & $-1.18$ {\scriptsize $[-1.87, -0.50]$} & $+3.44$ {\scriptsize $[+1.74, +5.23]$} \\
\bottomrule\end{tabular}\endgroup
\par\end{table}

\textbf{\phantomsection\label{app:C.7}C.7 MATH-500.} AIME25 and AIME26 contain 30 problems
each. MATH-500 \citep{arxiv210303874,arxiv230520050}
provides a larger mathematics test, evaluated with the same
caps and sampling but outside the six-task Total. At 16K,
Label is indistinguishable from the initial student at every
size, whereas DN-MOPD is ahead of Label at every size and
cap with every interval above zero. The larger models are
close to the ceiling on this benchmark, so the absolute
differences at 9B and 4B are small.

\begin{table}[!htbp]\centering
\definecolor{qgray}{RGB}{244,245,246}
\definecolor{qblue}{RGB}{233,242,249}
\caption{\textbf{MATH-500.} Mean accuracy (\%) over 500 problems with 16 samples each, graded by Math-Verify, and paired differences (pp) with 95\% question-bootstrap intervals. Students use 80 updates. This benchmark is not part of the six-task Total.}\label{tab:math500}
\begingroup\fontsize{9}{11}\selectfont\setlength{\tabcolsep}{5pt}\renewcommand{\arraystretch}{1.1}
\begin{tabular}{@{}lrrrrrr@{}}\toprule
Size & Initial & Label & Math-only OPD & DN-MOPD & Label $-$ initial & DN-MOPD $-$ Label \\\midrule
\addlinespace[2pt]\rowcolor{qgray}\multicolumn{7}{@{}l@{}}{\textit{16K evaluation}} \\
\hspace*{1em}9B & 95.84 & 95.76 & 97.30 & 96.50 & $-0.08$ {\scriptsize $[-0.51, +0.35]$} & $+0.74$ {\scriptsize $[+0.31, +1.18]$} \\
\hspace*{1em}4B & 94.46 & 94.08 & 95.84 & 95.33 & $-0.39$ {\scriptsize $[-0.91, +0.12]$} & $+1.25$ {\scriptsize $[+0.67, +1.85]$} \\
\hspace*{1em}2B & 76.91 & 77.21 & 82.93 & 83.16 & $+0.30$ {\scriptsize $[-0.64, +1.22]$} & $+5.95$ {\scriptsize $[+4.95, +6.96]$} \\
\addlinespace[2pt]\rowcolor{qgray}\multicolumn{7}{@{}l@{}}{\textit{8K evaluation}} \\
\hspace*{1em}9B & 94.08 & 94.38 & 96.80 & 95.83 & $+0.30$ {\scriptsize $[-0.19, +0.81]$} & $+1.45$ {\scriptsize $[+0.92, +2.00]$} \\
\hspace*{1em}4B & 92.01 & 92.89 & 95.24 & 94.55 & $+0.88$ {\scriptsize $[+0.31, +1.45]$} & $+1.66$ {\scriptsize $[+1.05, +2.29]$} \\
\hspace*{1em}2B & 68.40 & 72.91 & 80.90 & 79.76 & $+4.51$ {\scriptsize $[+3.54, +5.52]$} & $+6.85$ {\scriptsize $[+5.73, +7.99]$} \\
\bottomrule\end{tabular}\endgroup
\par\end{table}

\clearpage

\section{Implementation and diagnostic
details}\label{app:D}\label{appendix-d-implementation-and-diagnostic-details}

\setcounter{table}{0}\renewcommand{\thetable}{D.\arabic{table}}\renewcommand{\theHtable}{D.\arabic{table}}

\textbf{\phantomsection\label{app:D.1}D.1 Domain scaling and policy optimization.} In both
DN-MOPD and Label, every domain label selects its
corresponding expert, and the current student generates all
training responses.

For each rollout batch, let \(m_{i,t}\) mark a valid
response token, and form
\(r_{i,t}=\ell^T_{i,t}-\ell^{\mathrm{roll}}_{i,t}\) wherever
\(m_{i,t}=1\). All valid tokens with domain \(d_i=d\) enter
the same domain statistic. The standard deviations use the
population convention (division by the number of tokens),
not the sample convention. The global statistic is computed
after concatenating all domains' valid token values; it is
not an average of their standard deviations. The multiplier
is
\(\operatorname{clip}(\sigma_{\mathrm{all}}/\sigma_d,0.25,4)\),
with multiplier one if either standard deviation is zero or
has fewer than two observations. The same factor multiplies
every advantage in a sample from that domain. Statistics and
factors are recomputed for each batch and recorded in the
run audit.

There are two evaluations of the pre-update student. The
rollout service provides cached log-probabilities for the
scale estimate. The actor recomputes the response-token
log-probabilities before optimization; these enter the
advantage \(A=\ell^T-\ell^{\mathrm{actor}}\) and the
denominator of the clipped policy ratio. The reported
configuration multiplies this actor-side advantage by the
domain factor. It does not replace actor log-probabilities
with cached rollout values. All factors and advantages are
detached before policy optimization, and response masks and
loss reduction are inherited from the Label control.

The complete clipped policy-gradient token term is \[
\mathcal L_{i,t}(\theta)=-\min\!\left\{
\rho_{i,t}(\theta)\widetilde A_{i,t},\;
\operatorname{clip}\bigl(\rho_{i,t}(\theta),1-\eta_-,1+\eta_+\bigr)\widetilde A_{i,t}
\right\}.
\] \[
\rho_{i,t}(\theta)=\frac{\pi_\theta(y_{i,t}\mid h_{i,t})}{\pi_u(y_{i,t}\mid h_{i,t})}.
\] Here \(\pi_u\) is the pre-update student, \(\eta_-\) and
\(\eta_+\) are the baseline policy-ratio clipping bounds,
and
\(\widetilde A_{i,t}=\operatorname{stopgrad}(w_{d_i}A_{i,t})\).
The actor-side \(A_{i,t}\) uses the recomputed pre-update
student log-probability. Prompt and padding positions are
excluded through the response mask; the Label baseline's
masking and loss reduction are preserved. These
implementation choices are shared in the matched comparison.
The response-token loss is averaged within each response and
then across responses. This response-level reduction differs
from the token-pooled statistics used to estimate the
scaling factors. The policy-ratio clipping bounds are
\(\eta_-=\eta_+=0.2\). The additional scaling operation uses
already available scores and does not itself require another
teacher forward pass.

For a nondegenerate domain whose ratio is not clipped, the
scaled rollout signal has population standard deviation
\(\operatorname{Std}(w_d r)=\sigma_{\mathrm{all}}\) on that
batch. This identity concerns the measured rollout
log-ratios. It neither normalizes each domain to unit
variance nor establishes equality of actor-side gradients.
Domain means are scaled rather than subtracted, and the
positive factor preserves each advantage's sign. The lower
and upper clips limit the adjustment when a scale ratio is
extreme.

\textbf{Relation to Open-MOPD.} Open-MOPD
\citep{arxiv260819098} sets per-domain loss weights from a
running estimate of each domain's mean reward magnitude, and
reports that inverting this rule, so that domains with
smaller rewards receive larger weights, creates an unstable
feedback loop. DN-MOPD differs in three respects: it scales
by the spread of the log-ratios rather than by their mean
magnitude, it recomputes the statistic on every batch
instead of carrying a running estimate, and it bounds each
multiplier to \([0.25,4]\) rather than \([0.05,20]\). All
DN-MOPD runs reported here, including the 160-update
continuations and the two additional student seeds,
completed training and improved on Label.

The implementation changes only the advantage computation of
the Label training code; a reference implementation, the
training code and the configurations are released at
\url{https://github.com/LiXin97/DN-MOPD}.

\textbf{\phantomsection\label{app:D.2}D.2 Feedback measurements.} \hyperref[fig:4]{Figure 4} (panel a)
reports the ratio of each domain's rollout log-ratio
standard deviation to the pooled standard deviation. Points
are medians and whiskers are interquartile ranges over the
recorded batches of the DN-MOPD runs, which include their
continuation beyond the 80-update endpoint: the logged
batches run through updates 115, 131 and 160 at 9B, 4B and
2B (134 batches at 4B, three of them repeats after a
restart), while the multiplier traces in panel (c) run
through updates 125, 144 and 160. Domain statistics pool
valid response tokens, so longer answers contribute more
observations than shorter ones.

The imbalance is present before normalization can affect the
student. At the first update, DN-MOPD samples from the same
initial student as Label, so its first batch measures the
feedback with which Label training also starts. There, the
spread ratios for mathematics, code and IF are 0.57, 1.13
and 4.41 at 9B, 0.50, 1.23 and 2.93 at 4B, and 0.48, 1.22
and 2.32 at 2B (Table \hyperref[app:D.1]{D.1}). The IF spread grows during
training, so the medians in \hyperref[fig:4]{Figure 4} (panel a) exceed these
starting values, while mathematics remains below the pooled
spread throughout.

\hyperref[fig:4]{Figure 4} (panel b) reports DN-MOPD minus Label in each
evaluation domain at 16K, computed from the same six-task
records as \hyperref[tab:main]{Tables 1} and \hyperref[tab:scales]{2}. \hyperref[fig:4]{Figure 4} (panel c) uses the last
recorded attempt for each rollout identifier and plots the
applied multiplier, together with the raw IF multiplier
before clipping. These traces include the training
continuations; their archived endpoints are updates 125, 144
and 160 at 9B, 4B and 2B. The dashed vertical line marks the
main 80-update endpoint. No missing continuation values are
extrapolated.

The batch summaries and trajectories describe the measured
feedback signal. They show a correspondence between
mathematics amplification and mathematics gains, but the
fixed-weight controls (\hyperref[app:C.6]{App. C.6}) show that reducing IF alone
produces most of the mathematics gain, so the correspondence
does not identify amplification as the cause. At 9B the code
multiplier is also raised, yet code does not improve at 16K
(\hyperref[app:C.3]{App. C.3}), so amplification alone does not guarantee a
domain gain. \hyperref[app:D.5]{App. D.5} measures each domain's contribution to
the shared parameter update directly.

\textbf{\phantomsection\label{app:D.3}D.3 Scale imbalance, clipping and token
composition.} Table \ref{tab:feedback:diagnostics} expands
the feedback summary with the raw multiplier, its applied
value after clipping, and each domain's share of response
tokens. Mathematics receives an increased multiplier across
all three pools. IF has a much larger log-ratio spread, so
its multiplier often reaches the lower clipping bound. The
clipping frequency therefore matters when interpreting
normalization: the applied rule limits IF feedback without
fully equalizing its spread.

\begin{table}[!htbp]\centering
\definecolor{qgray}{RGB}{244,245,246}
\definecolor{qblue}{RGB}{233,242,249}
\caption{\textbf{Feedback scale, clipping and token composition.} The first column is the spread ratio in the first training batch, sampled from the same initial student as Label. The other columns are medians over the recorded batches of the DN-MOPD runs, including their continuation beyond 80 updates (\hyperref[app:D.2]{App. D.2}); the clipping column counts logged batches where the factor reaches a bound. Token share is the within-batch fraction of valid response tokens.}\label{tab:feedback:diagnostics}
\begingroup\fontsize{9}{11}\selectfont\setlength{\tabcolsep}{5pt}\renewcommand{\arraystretch}{1.1}
\begin{tabular}{@{}lrrrrrr@{}}\toprule
Domain & Update 0 & $\sigma_d/\sigma_{\mathrm{all}}$ & Raw factor & Applied factor & Clipped & Tokens (\%) \\\midrule
\addlinespace[2pt]\rowcolor{qgray}\multicolumn{7}{@{}l@{}}{\textit{Qwen3.5-9B}} \\
\hspace*{1em}Math & 0.57 & 0.65 & 1.54 & 1.54 & 0/115 & 48.9 \\
\hspace*{1em}Code & 1.13 & 0.64 & 1.57 & 1.57 & 0/115 & 50.1 \\
\hspace*{1em}IF & 4.41 & 7.37 & 0.14 & 0.25 & 115/115 & 0.9 \\
\addlinespace[2pt]\rowcolor{qgray}\multicolumn{7}{@{}l@{}}{\textit{Qwen3.5-4B}} \\
\hspace*{1em}Math & 0.50 & 0.49 & 2.03 & 2.03 & 0/134 & 46.9 \\
\hspace*{1em}Code & 1.23 & 1.02 & 0.98 & 0.98 & 0/134 & 52.5 \\
\hspace*{1em}IF & 2.93 & 6.68 & 0.15 & 0.25 & 99/134 & 0.7 \\
\addlinespace[2pt]\rowcolor{qgray}\multicolumn{7}{@{}l@{}}{\textit{Qwen3.5-2B}} \\
\hspace*{1em}Math & 0.48 & 0.49 & 2.06 & 2.06 & 0/160 & 46.4 \\
\hspace*{1em}Code & 1.22 & 1.12 & 0.89 & 0.89 & 0/160 & 52.9 \\
\hspace*{1em}IF & 2.32 & 5.85 & 0.17 & 0.25 & 135/160 & 0.6 \\
\bottomrule\end{tabular}\endgroup
\par\end{table}

The pooled variance includes both within-domain dispersion
and differences in domain means. With \(p_d\) denoting a
domain's fraction of valid response tokens and \(\mu_d\) its
mean log-ratio, \[
\sigma_{\mathrm{all}}^2=\sum_d p_d\left[\sigma_d^2+(\mu_d-\mu_{\mathrm{all}})^2\right].
\] Consequently, the common target scale depends on the
batch's domain composition as well as the individual
spreads. DN-MOPD multiplies the original advantages; it does
not subtract domain means. This also explains why the rule
should not be interpreted as independent unit-variance
normalization of each domain.

IF contributes few response tokens because its answers are
short. This does not imply that IF has a negligible
contribution to the training loss: token shares describe the
scale estimator, whereas the loss averages within responses
before averaging across responses. Nor does the multiplier
alone determine the resulting gradient. The different code
outcomes across sizes are a concrete reason to distinguish
feedback scaling from an established account of
parameter-update interference; \hyperref[app:D.5]{App. D.5} reports the
per-domain gradients at 4B.

\textbf{\phantomsection\label{app:D.4}D.4 What the controls establish.} The Label
comparison holds teacher identities fixed and changes the
scale applied to their feedback. Uniform pooling and Dynamic
examine alternative teacher rules; annealed injection
retains Label assignments but adds an early imitation
signal. Together these comparisons support feedback scaling
as a useful change to the tested Label recipe. The
fixed-weight controls (\hyperref[app:C.6]{App. C.6}) add that constant weights
measured before training recover the benefit at 9B and 4B,
while per-batch estimation adds to it at 2B. The experiments
do not separately vary the clipping bounds, domain
proportions or the statistic used to estimate scale.

\textbf{\phantomsection\label{app:D.5}D.5 Sources of the pooled spread and of the update.}
IF answers are short, so IF contributes about 1\% of the
valid response tokens, but its large spread gives it 51\%,
21\% and 19\% of the pooled log-ratio variance at 9B, 4B and
2B; differences between domain means contribute about 1\%.
The imbalance is not specific to DN-MOPD training. The
seed-42 to seed-44 Label runs record, without applying, the
multiplier DN-MOPD would use; these stay near 1.5--1.9 for
mathematics and 0.25--0.34 for IF throughout training, with
almost no variation across seeds.

\begin{table}[!htbp]\centering
\definecolor{qgray}{RGB}{244,245,246}
\definecolor{qblue}{RGB}{233,242,249}
\caption{\textbf{Sources of the pooled spread.} Token share and each domain\textquotesingle s share of the pooled rollout log-ratio variance (within-domain plus mean-difference terms), averaged over updates 0--79 of the DN-MOPD runs. The last two columns give the multiplier DN-MOPD would apply, recorded but not applied during three Label runs (range over seeds of the mean over updates 0--79), and the multiplier DN-MOPD applied.}\label{tab:scale:sources}
\begingroup\fontsize{9}{11}\selectfont\setlength{\tabcolsep}{5pt}\renewcommand{\arraystretch}{1.1}
\begin{tabular}{@{}lrrrr@{}}\toprule
Domain & Tokens (\%) & Pooled variance (\%) & Would-be under Label & Applied by DN-MOPD \\\midrule
\addlinespace[2pt]\rowcolor{qgray}\multicolumn{5}{@{}l@{}}{\textit{Qwen3.5-9B}} \\
\hspace*{1em}Math & 48.4 & 20.8 & 1.49 & 1.59 \\
\hspace*{1em}Code & 50.6 & 28.3 & 0.99--1.00 & 1.53 \\
\hspace*{1em}IF & 1.0 & 50.9 & 0.25 & 0.25 \\
\addlinespace[2pt]\rowcolor{qgray}\multicolumn{5}{@{}l@{}}{\textit{Qwen3.5-4B}} \\
\hspace*{1em}Math & 47.2 & 10.9 & 1.91--1.93 & 2.15 \\
\hspace*{1em}Code & 52.0 & 67.9 & 0.79 & 0.89 \\
\hspace*{1em}IF & 0.8 & 21.2 & 0.34 & 0.28 \\
\addlinespace[2pt]\rowcolor{qgray}\multicolumn{5}{@{}l@{}}{\textit{Qwen3.5-2B}} \\
\hspace*{1em}Math & 46.5 & 10.3 & 1.76--1.78 & 2.18 \\
\hspace*{1em}Code & 52.8 & 70.9 & 0.81 & 0.86 \\
\hspace*{1em}IF & 0.7 & 18.8 & 0.31 & 0.28 \\
\bottomrule\end{tabular}\endgroup
\par\end{table}

To measure contributions to the shared update, we computed
the gradient of each domain's OPD loss over all trainable
parameters of the 4B student, with 32 prompts per domain,
four sampled responses each and the response-averaged
reduction used in training. At the initial student, the IF
gradient norm is 5 to 15 times those of code and
mathematics, and IF supplies 94\% of the combined gradient
under equal weights. Under DN-MOPD's first-batch weights its
share falls to 64\%, while mathematics and code rise from
1\% and 5\% to 16\% and 20\%. The IF gradient is also the
least consistent: the cosine between the gradients of two
disjoint prompt halves is 0.41 for IF against 0.89 and 0.88
for mathematics and code, and it falls to \(-0.07\) at the
Label checkpoint after 80 updates, where a few very short IF
answers carry most of the IF gradient. After training, IF
dominates the combined gradient under either weighting.
Averaging the loss over tokens instead of responses does not
remove the imbalance at the initial student: IF would still
supply 88\% of the combined gradient under equal weights and
45\% under DN-MOPD's first-batch weights. The residual
feedback also differs after training. On the probe answers,
the standard deviation of the mathematics and code
log-ratios falls from 0.31 and 0.68 at initialization to
0.20 and 0.41 for Label after 80 updates, but to 0.07 and
0.12 for DN-MOPD; the IF values stay between 1.1 and 1.7.
This is consistent with the DN-MOPD student having absorbed
more of the mathematics and code teachers' feedback, so that
little gradient remains in these domains and IF stays
dominant. These measurements are descriptive: they use one
probe set, gradients at policy ratio one without clipping or
optimizer state, and they do not by themselves establish the
cause of the capability differences.

\begin{table}[!htbp]\centering
\definecolor{qgray}{RGB}{244,245,246}
\definecolor{qblue}{RGB}{233,242,249}
\caption{\textbf{Per-domain gradients of the OPD loss at 4B.} Gradients over all trainable parameters for 32 prompts per domain and four sampled responses each, using the response-averaged loss of training. Split-half cosine compares the gradients of two disjoint prompt halves. Shares are each domain\textquotesingle s contribution $\langle w_d g_d, g\rangle/\lVert g\rVert^2$ to the combined gradient $g=\sum_d w_d g_d$, under equal weights and under DN-MOPD\textquotesingle s first-batch weights (2.01 / 0.81 / 0.34). Descriptive, one probe set; a replicate draw preserves the ordering.}\label{tab:gradients}
\begingroup\fontsize{9}{11}\selectfont\setlength{\tabcolsep}{5pt}\renewcommand{\arraystretch}{1.1}
\begin{tabular}{@{}lrrrr@{}}\toprule
Domain & $\lVert g_d\rVert$ & Split-half cos. & Share, equal (\%) & Share, DN (\%) \\\midrule
\addlinespace[2pt]\rowcolor{qgray}\multicolumn{5}{@{}l@{}}{\textit{Initial student}} \\
\hspace*{1em}Math & 3.6 & $0.89$ & 1 & 16 \\
\hspace*{1em}Code & 10.6 & $0.88$ & 5 & 20 \\
\hspace*{1em}IF & 51.7 & $0.41$ & 94 & 64 \\
\addlinespace[2pt]\rowcolor{qgray}\multicolumn{5}{@{}l@{}}{\textit{Label, 80 updates}} \\
\hspace*{1em}Math & 2.6 & $0.90$ & 0 & 0 \\
\hspace*{1em}Code & 10.2 & $0.70$ & 0 & 1 \\
\hspace*{1em}IF & 315.6 & $-0.07$ & 100 & 99 \\
\addlinespace[2pt]\rowcolor{qgray}\multicolumn{5}{@{}l@{}}{\textit{DN-MOPD, 80 updates}} \\
\hspace*{1em}Math & 0.7 & $0.74$ & 0 & 0 \\
\hspace*{1em}Code & 2.4 & $0.57$ & 0 & 0 \\
\hspace*{1em}IF & 223.5 & $0.21$ & 100 & 100 \\
\bottomrule\end{tabular}\endgroup
\par\end{table}

\section{Capability transfer and a boundary
case}\label{app:E}\label{appendix-e-capability-transfer-and-a-boundary-case}

\setcounter{table}{0}\renewcommand{\thetable}{E.\arabic{table}}\renewcommand{\theHtable}{E.\arabic{table}}

\textbf{\phantomsection\label{app:E.1}E.1 Capability transfer in Qwen3.5.} \hyperref[fig:3]{Figure 3}
compares each expert with the shared initialization across
evaluation domains. Every specialist improves its own domain
at every size, while its effects on other domains vary.
Label nevertheless retains little of the mathematics
specialist's gain. Mathematics-only OPD preserves much more
of that gain under the same update budget, showing that the
student can learn from this teacher. The loss of transfer
appears when its feedback is combined with the other
domains.

We measure retained mathematics gain as the student's
improvement over initialization divided by the mathematics
expert's improvement over the same initialization. At 8K,
DN-MOPD retains 61--90\% of that gain across sizes, compared
with 14--32\% for Label. Table \ref{tab:math:transfer}
reports the absolute mathematics gains at both caps with
paired 95\% intervals. At 16K, Label shows no mathematics
gain over initialization at any size and falls below it at
4B, whereas DN-MOPD improves mathematics at every size. At
4B, neither the specialist nor math-only OPD gains separably
at 16K, so retained fractions are reported at 8K only. Using
absolute gains alongside retained fractions avoids hiding
this loss behind an aggregate Total. The ratio is sensitive
to the expert's gain over initialization, so it is a
within-construction diagnostic rather than a
scale-independent measure of distillation quality.

\begin{table}[!htbp]\centering
\definecolor{qgray}{RGB}{244,245,246}
\definecolor{qblue}{RGB}{233,242,249}
\caption{\textbf{Mathematics capability transferred from the expert.} Mathematics gains over the shared initialization (pp) with paired 95\% question-bootstrap intervals. The single-teacher student learns only from the mathematics expert. Students use 80 updates; all four gains in a row share the same initialization and evaluation cap.}\label{tab:math:transfer}
\begingroup\fontsize{9}{11}\selectfont\setlength{\tabcolsep}{5pt}\renewcommand{\arraystretch}{1.1}
\begin{tabular}{@{}lrrrr@{}}\toprule
Size & Math expert & Math-only OPD & Label & DN-MOPD \\\midrule
\addlinespace[2pt]\rowcolor{qgray}\multicolumn{5}{@{}l@{}}{\textit{Evaluation cap: 16K}} \\
\hspace*{1em}9B & $+4.45$ {\scriptsize $[+2.55, +6.35]$} & $+4.09$ {\scriptsize $[+2.16, +6.04]$} & $-0.86$ {\scriptsize $[-2.42, +0.76]$} & $+3.12$ {\scriptsize $[+1.20, +5.03]$} \\
\hspace*{1em}4B & $+2.06$ {\scriptsize $[-0.23, +4.53]$} & $+1.69$ {\scriptsize $[-0.42, +3.91]$} & $-1.98$ {\scriptsize $[-3.78, -0.18]$} & $+1.77$ {\scriptsize $[+0.05, +3.65]$} \\
\hspace*{1em}2B & $+4.58$ {\scriptsize $[+2.47, +6.95]$} & $+4.51$ {\scriptsize $[+2.45, +6.74]$} & $-0.60$ {\scriptsize $[-1.88, +0.73]$} & $+3.12$ {\scriptsize $[+1.25, +5.26]$} \\
\addlinespace[2pt]\rowcolor{qgray}\multicolumn{5}{@{}l@{}}{\textit{Evaluation cap: 8K}} \\
\hspace*{1em}9B & $+12.32$ {\scriptsize $[+8.93, +15.76]$} & $+10.70$ {\scriptsize $[+7.50, +14.04]$} & $+3.05$ {\scriptsize $[+1.48, +4.66]$} & $+7.55$ {\scriptsize $[+5.26, +9.95]$} \\
\hspace*{1em}4B & $+12.86$ {\scriptsize $[+9.24, +16.77]$} & $+11.74$ {\scriptsize $[+8.12, +15.50]$} & $+1.82$ {\scriptsize $[+0.08, +3.65]$} & $+9.24$ {\scriptsize $[+6.25, +12.34]$} \\
\hspace*{1em}2B & $+11.02$ {\scriptsize $[+7.08, +15.13]$} & $+10.62$ {\scriptsize $[+7.14, +14.32]$} & $+3.57$ {\scriptsize $[+2.19, +5.08]$} & $+9.92$ {\scriptsize $[+6.64, +13.44]$} \\
\bottomrule\end{tabular}\endgroup
\par\end{table}

\textbf{\phantomsection\label{app:E.2}E.2 Earlier Qwen3-4B comparison.} The earlier
construction uses a Qwen3-4B student \citep{arxiv250509388},
three same-base experts after 400 RL updates, and the
2,700-prompt domain mixture. Label and DN-MOPD are compared
at 160 student updates with seed 42 and a 16,384-token
training response cap. Their public evaluation uses the same
six-suite averaging definition and a 16,384-token response
cap. This differs from the Qwen3.5 main comparison in model
family, expert checkpoints, training cap and training
duration.

\begin{table}[!htbp]\centering
\definecolor{qgray}{RGB}{244,245,246}
\definecolor{qblue}{RGB}{233,242,249}
\caption{\textbf{Earlier Qwen3-4B boundary case.} Domain means and six-task Total (\%) at 160 updates and a 16K evaluation cap. Differences are DN-MOPD minus Label (pp), with paired 95\% question-bootstrap intervals. The intervals include zero for every domain and Total.}\label{tab:dn:qwen3boundary}
\begingroup\fontsize{9}{11}\selectfont\setlength{\tabcolsep}{5pt}\renewcommand{\arraystretch}{1.1}
\begin{tabular}{@{}lrrrr@{}}\toprule
Metric & Label & DN-MOPD & $\Delta$ (pp) & 95\% CI \\\midrule
Math & 53.67 & 53.41 & $-0.26$ & $[-1.35, +0.76]$ \\
Code & 38.36 & 38.27 & $-0.08$ & $[-1.32, +1.16]$ \\
IF & 56.13 & 56.53 & $+0.40$ & $[-0.19, +1.04]$ \\
\midrule
\rowcolor{qblue}Total & 49.39 & 49.41 & $+0.02$ & $[-0.56, +0.59]$ \\
\bottomrule\end{tabular}\endgroup
\par\end{table}

Label already transfers substantial mathematics capability
in this construction, and DN-MOPD leaves the mathematics
signal near its original scale: the median applied
multiplier is 1.00 for mathematics, 1.22 for code and 0.37
for IF (clipped in 29 of 172 logged batches), against
1.54--2.06 for mathematics in the Qwen3.5 runs (Table \hyperref[app:D.1]{D.1}).
The domain-resolved results show no clear improvement in
mathematics or code; the small IF increase is also
uncertain. This supports a configuration-dependent benefit,
but model and protocol differences prevent treating the two
constructions as a controlled test of imbalance.

\end{document}